\documentclass[sigconf]{acmart}

\renewcommand\footnotetextcopyrightpermission[1]{}

\usepackage{amsmath}
\usepackage{algorithm}
\usepackage{algorithmic}
\usepackage{amsfonts}
\usepackage{graphicx}
\usepackage{booktabs}
\usepackage{enumitem}
\usepackage{bm}
\usepackage{dsfont}
\usepackage{multirow}
\usepackage{subfigure}
\usepackage{diagbox}
\usepackage{pifont}
\usepackage{xcolor}
\usepackage{xspace}
\usepackage{colortbl}
\usepackage{listings}

\AtBeginDocument{%
  }

\copyrightyear{2027}
\acmYear{2027}
\setcopyright{cc}
\setcctype{by}
\acmConference[Conference]
{}{2027}{USA}

\newcommand{\m}{TrAC\xspace}
\newcommand{\pce}{Prefix-Conditioned Elicitation\xspace}
\newcommand{\pcem}{PCE\xspace}
\newcommand{\tup}{Trace Uncertainty Profile\xspace}
\newcommand{\tupm}{TUP\xspace}
\newcommand{\cf}{Consensus Fusion\xspace}
\newcommand{\cfm}{CF\xspace}

\newcommand\blue[1]{\textcolor{blue}{#1}}

\begin{document}

\title[\m]{\m: Trace-Conditioned Answer Consistency for Efficient Uncertainty Quantification in LLMs}

\author{Dahai Yu}
\affiliation{
  \institution{Florida State University}
  \city{Tallahassee, Florida}
  \country{USA}
}
\email{dahai.yu@fsu.edu}

\author{Lin Jiang}
\affiliation{
  \institution{Florida State University}
  \city{Tallahassee, Florida}
  \country{USA}
}
\email{lin.jiang@fsu.edu}

\author{Rongchao Xu}
\affiliation{
  \institution{Florida State University}
  \city{Tallahassee, Florida}
  \country{USA}
}
\email{rxu@fsu.edu}

\author{Guang Wang}
\authornote{Prof. Guang Wang is the corresponding author.}
\affiliation{
  \institution{Florida State University}
  \city{Tallahassee, Florida}
  \country{USA}
}
\email{guang@cs.fsu.edu}

\begin{abstract}
Large language models (LLMs) can generate fluent reasoning traces that nevertheless lead to incorrect answers, making response-level uncertainty estimation important for abstention, human review, and adaptive compute allocation. 
Existing approaches generally fall into three categories: passive single-trace methods use token-level confidence signals, sampling-based methods compare multiple complete traces at higher generation cost, and active prefix-based methods probe partial traces to study answer stabilization or preference transitions. However, none actively re-elicits an answer from a completed reasoning trace to measure its consistency with and support for the original answer.
To address this gap, we introduce Trace-Conditioned Answer Consistency (\m), a correctness-supervised uncertainty quantification framework that combines active and passive signals anchored to one completed reasoning trace. Its active component, \textbf{\pce (\pcem)}, re-elicits a short answer conditioned on the completed trace and represents both its consistency with the original answer and its token-level probabilistic support. Its passive component, \textbf{\tup (\tupm)}, summarizes how token-level uncertainty evolves throughout the original generation without additional decoding. A lightweight head then integrates the two representations into a response-correctness score.
Across five mathematical reasoning benchmarks and three LLM families, \m improves macro AUROC by 1.8\% and reduces AURC by 3.4\% relative to eight-sample self-consistency, while using one complete reasoning trace and a short cached answer probe. When eight samples are already available, augmenting sample consensus with re-elicitation further improves macro AUROC by 4.3\% and reduces AURC by 8.3\%, without additional full-trace generation. These results establish answer re-elicitation as a low-overhead uncertainty signal that remains informative even when passive confidence is misleading or multiple samples converge on an incorrect answer.
\end{abstract}



\maketitle

\section{Introduction}\label{sec:introduction}

Large language models (LLMs) solve complex problems by generating long chains of reasoning~\cite{wei2022cot,kojima2022zeroshot}, yet a fluent derivation can still end in an incorrect answer. A deployment system therefore needs a response-level uncertainty score that ranks reliable outputs ahead of likely errors.  Such a score supports selective prediction, routes uncertain cases to human review, and determines when an expensive second round of sampling is warranted~\cite{geifman2017selective}. We study this problem in verifiable mathematical reasoning, where deterministic answer checking provides clean offline labels, while confident text still need not imply a correct conclusion~\cite{kadavath2022mostly,vashurin2025polygraph}. Our goal is not to invoke a separate judge or generate another full reasoning trace, but to estimate the correctness of one generated response using signals available at inference time.

Existing uncertainty estimators largely fall into three categories. \textit{Passive single-trace methods} use token-level confidence signals already available from a completed response~\cite{kang2025selfcertainty,fu2025deepconf}, but do not actively test whether the completed reasoning context continues to support the returned answer. \textit{Vote-based sampling methods} estimate confidence from agreement across multiple independently generated traces~\cite{wang2023selfconsistency,taubenfeld2025cisc}, but require several full generations and lose ranking resolution when all sampled answers agree. \textit{Active prefix-based methods} elicit answers from partial reasoning prefixes to study stabilization or preference transitions, primarily for early stopping rather than the correctness of a completed response. This leaves a key gap: an active answer-level probe that re-elicits an answer from a completed trace without generating another full reasoning trace.

Our key intuition is that re-eliciting an answer from the completed reasoning prefix can reveal whether, and how strongly, the model continues to support its original answer. Specifically, we append a short, fixed cue to the completed trace and re-elicit only an answer suffix from the same frozen model. If the trace robustly supports its conclusion, the re-elicited answer should reproduce the original answer with high likelihood. If the conclusion is unstable, the re-elicited answer may disagree or receive weak probabilistic support. We refer to the measured property as \emph{\textbf{trace-conditioned answer consistency}} and to the procedure used to measure it as \emph{\textbf{answer re-elicitation}}. Unlike explicit self-verification, which asks the model to judge whether its answer is correct, answer re-elicitation measures answer reproducibility under a controlled re-reading of the completed trace. It also differs from cross-trace consensus: rather than comparing conclusions reached through different reasoning traces, it probes answer support within a single completed trace. Even its simplest signal is strongly informative: the original and re-elicited answers agree far more often on correct traces than on incorrect ones.

However, turning this observation into an effective uncertainty estimator presents \textbf{\textit{three challenges}}. First, the answer should be re-elicited at low cost without a judge model, a reference answer, or another complete reasoning trace. Second, open-form answers of variable token length require a fixed representation that preserves both answer identity and strength of support. Third, it is important to determine whether active re-elicitation provides information beyond passive token confidence and the model's own self-verification probability.

To address these challenges, we introduce Trace-Conditioned Answer Consistency (\m), a correctness-supervised uncertainty quantification (UQ) framework that combines two complementary views derived from one complete reasoning trace. First, \textbf{\pce (\pcem)} re-elicits an answer from the completed prefix and encodes its agreement, likelihood, and confidence relative to the returned answer. Second, the \textbf{\tup (\tupm)} summarizes token-level uncertainty already available from the original generation, requiring no additional decoding. A lightweight prediction head is then designed to integrate the active re-elicitation and passive trace-profile representations into a response-correctness score.

We evaluate \m on five mathematical reasoning benchmarks and three LLM families. Extensive experiments show that \m improves macro AUROC by 1.8\% and reduces AURC by 3.4\% compared with eight-sample self-consistency, while the short answer probe adds only 2\% measured latency. When eight samples are already available, augmenting sample consensus with re-elicitation further improves macro AUROC by 4.3\% and reduces AURC by 8.3\% at the same full-generation budget. Controlled ablations show that active re-elicitation and passive trace-profile evidence are complementary. They further demonstrate that re-elicitation captures information distinct from explicit self-verification and remains informative under greedy decoding.

The key contributions of this work are as follows:
\begin{itemize}

\item \textbf{Conceptually}, we formulate trace-conditioned answer consistency as a new active uncertainty signal, measured by re-eliciting an answer from one completed reasoning trace and testing whether the model still returns the answer it originally produced.

\item \textbf{Technically}, we propose \m, a correctness-supervised UQ framework that combines two complementary active and passive views derived from one completed reasoning trace: \pce (\pcem), which provides a low-cost representation of open-form answer re-elicitation, and the \tup (\tupm), which summarizes token-level uncertainty from the original trace.

\item \textbf{Empirically}, we evaluate \m across five benchmarks and three LLM families. Extensive experiments show that using one complete reasoning trace, \m outperforms eight-sample self-consistency. When multiple traces are available, it further strengthens consensus-based uncertainty estimation without additional full-trace generation. Code is available at 
\blue{\url{https://github.com/UFOdestiny/TrAC}}.

\end{itemize}

\section{Problem Formulation}\label{sec:preliminary}

We study response-level UQ for reasoning language models. Given a reasoning problem \(x\), a frozen language model \(p_{\theta}\) generates a response
\begin{equation}
y = (r,a),
\end{equation}
where \(r\) denotes the completed reasoning trace and \(a\) is the final answer returned by the model. Our goal is to estimate whether this particular response is correct, rather than to generate a new answer.
During offline training and evaluation, a deterministic verifier \(V\) compares the returned answer \(a\) with a reference answer \(a^{\ast}\) and provides a binary correctness label
\begin{equation}
z = V(a,a^{\ast}) \in \{0,1\},
\end{equation}
where \(z=1\) indicates that the returned answer is correct. The reference answer \(a^{\ast}\) and the verifier \(V\) are used only to construct supervision labels and are not available to the uncertainty estimator at inference time.

The UQ task is to learn a scalar scoring function
\begin{equation}
u(x,y) \in [0,1],
\end{equation}
where \(u(x,y)\) estimates the reliability of a generated response \(y\) to an input \(x\). Although the estimator assigns one score to each response, these scores are compared across different responses to rank them by their likelihood of being correct. For example, if two responses receive scores \(0.8\) and \(0.3\), the former should be more likely to be correct. Formally, let \((x^{+},y^{+})\) and \((x^{-},y^{-})\) denote a randomly selected correct and an incorrect response, respectively. A desirable estimator should satisfy
\begin{equation}
u(x^{+},y^{+}) > u(x^{-},y^{-})
\end{equation}
as often as possible. Given a set of questions and their generated responses, the ranking quality of \(u\) can be characterized by the area under the receiver operating characteristic curve (AUROC)~\cite{gruenefeld2026tracing, pandey2026selfdoubt}, and more details will be in Section \ref{sec:metrics}.
At inference time, the estimator may access the problem \(x\), the generated response \(y=(r,a)\), and model-derived observations obtained from the frozen language model. It cannot access the reference answer \(a^{\ast}\), the verifier \(V\), or the correctness label \(z\). In this work, we introduce an auxiliary re-elicited answer \(\tilde{a}\), which is generated from the completed reasoning context and used only as an inference-time uncertainty signal.
\section{Methodology}\label{sec:method}

\begin{figure*}[t]
  \centering
  \includegraphics[width=\textwidth]{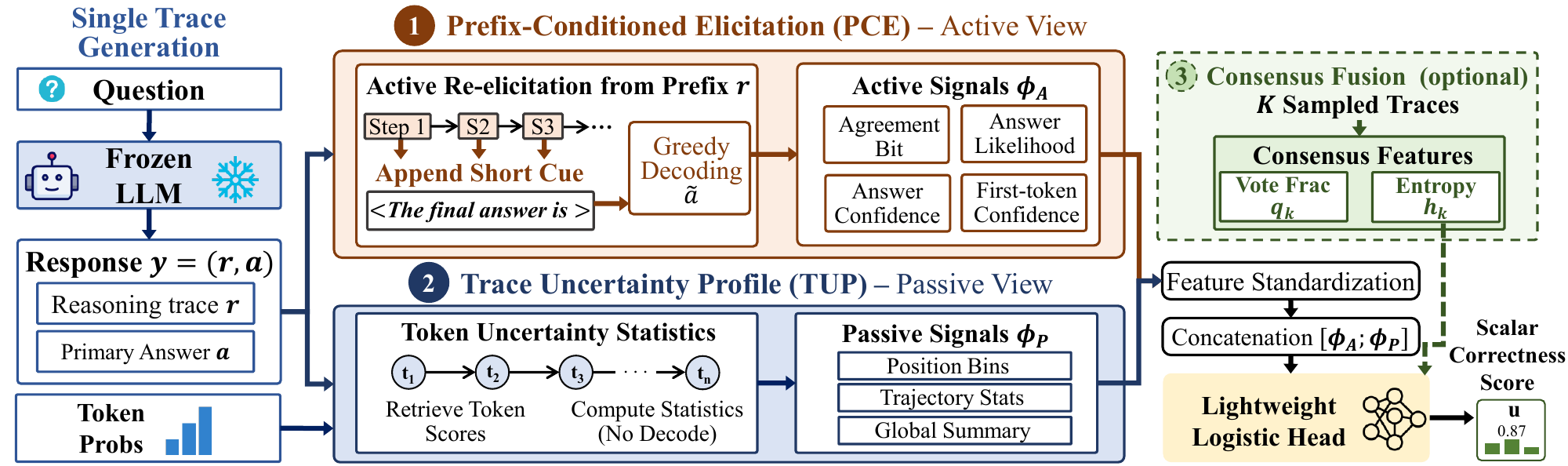}
  \caption{Overall framework of \m. \pcem re-elicits a short answer at the completed reasoning prefix and encodes its agreement together with answer likelihood and confidence. \tupm independently summarizes token uncertainty along the completed trace with no extra decoding. A shared logistic head fuses the two representations into one single-trace score. When several completed samples are already available, the optional consensus fusion provides additional vote statistics. }
  \Description{}
  \label{fig:framework}
\end{figure*}

As shown in Figure~\ref{fig:framework}, \m scores a completed reasoning trace by combining two complementary views anchored to that trace. The active \pcem module re-elicits a short answer from the completed reasoning prefix and measures how consistently the model reproduces its original answer. The passive \tupm module summarizes token-level uncertainty already available from the original generation and requires no additional decoding. A lightweight prediction head integrates the two representations into a response-correctness score. When multiple traces are already available, an optional consensus fusion module further incorporates vote-based statistics to improve correctness estimation.
Algorithm~\ref{alg:trac} summarizes the complete scoring pipeline, and \textbf{Appendix~\ref{app:example} provides an example} illustrating how the active and passive signals are combined to distinguish a supported answer from an inconsistent one.

\subsection{\pce (\pcem)}
\label{sec:pce}

\subsubsection{Prefix-Conditioned Re-Elicitation}
The first challenge is to obtain answer-level evidence from one completed trace without generating another reasoning trajectory. The completed response contains a primary answer $a$, but this realized output alone does not reveal how firmly the completed reasoning context supports it. Given the problem $x$, completed reasoning body $r$, and primary answer $a$, \pcem performs a controlled answer-only re-elicitation and constructs a fixed representation of how consistently the model reproduces $a$ and how strongly it supports the reproduced answer.

We define a fixed re-elicitation operator
\begin{equation}
(\tilde a,\mathbf c)=Q_\theta(x,r),
\label{eq:reelicit_operator}
\end{equation}
where $\mathbf c$ contains the token log-probabilities of the re-elicited answer. The operator reuses the completed reasoning body $r$ and appends a model-compatible end-of-thinking marker followed by the fixed cue
\texttt{$\langle$The final answer is$\rangle$}.
Greedy decoding then produces $\tilde a$ and $\mathbf c$. The cue is fixed across all examples, while the model parameters and original reasoning body remain unchanged. With prefix caching, the procedure decodes only a short answer suffix rather than another complete reasoning trace.
The resulting $\tilde a$ provides a controlled second elicitation from the same completed reasoning context. When the trace robustly supports its conclusion, $\tilde a$ is likely to reproduce $a$ with strong probabilistic support. When the conclusion is unstable, the two answers may disagree, or the re-elicited answer may receive weak support.

\subsubsection{Answer Consistency}
The primary signal is whether the re-elicited answer matches the originally returned answer:
\begin{equation}
e=\mathbb{I}[\tilde a\equiv a],
\end{equation}
where $\equiv$ applies deterministic answer normalization and equivalence checking without access to the reference answer. Agreement is strongly associated with correctness: across our evaluation matrix, the primary and re-elicited answers agree on 87.6\% of correct traces but only 24.7\% of incorrect traces. A single cached answer decode is sufficient to obtain this signal.

\subsubsection{Answer Likelihood and Confidence}

Because agreement provides only a binary observation, \pcem additionally measures the probabilistic support for the re-elicited answer. Let 
$\tilde a=(v_1,\ldots,v_L)$ denote its decoded tokens. We compute the length-normalized answer log-likelihood and minimum token log-likelihood:

\begin{equation}
\begin{aligned}
\ell &= \frac{1}{L}\sum_{i=1}^{L}\log p_\theta(v_i\mid x,r,\mathrm{cue},v_{<i}),
\end{aligned}
\label{eq:answer_likelihood}
\end{equation}

\begin{equation}
\begin{aligned}
\ell_{\min} &= \min_i \log p_\theta(v_i\mid x,r,\mathrm{cue},v_{<i}).
\end{aligned}
\label{eq:answer_likelihoodmin}
\end{equation}

We also define the answer-head log-likelihood over the first
$m=\min(2,L)$ tokens,
\begin{equation}
\ell_{\mathrm{head}}=
\frac{1}{m}\sum_{i=1}^{m}
\log p_\theta(v_i\mid x,r,\mathrm{cue},v_{<i}),
\end{equation}
and the first-token log-confidence
\begin{equation}
c_1=\log p_\theta(v_1\mid x,r,\mathrm{cue}).
\end{equation}

Length normalization makes likelihoods more comparable across answers with different token lengths, while $\ell_{\min}$ captures the least-supported answer token. The head statistic emphasizes confidence at the onset of the re-elicited answer, and $c_1$ remains available even when complete answer parsing fails. The active representation concatenates agreement, likelihood, and confidence:

\begin{equation}
\phi_A=
[\,e,\ \ell,\ \ell_{\min},\ \ell_{\mathrm{head}},\ c_1\,].
\label{eq:active_representation}
\end{equation}
Then we obtain the \pcem estimator
\begin{equation}
u_{\mathrm{active}}
=
\sigma(\mathbf w_A^\top\phi_A+b_A).
\label{eq:active_score}
\end{equation}
It captures both whether the model reproduces its original answer and how strongly it supports the re-elicited answer. These two aspects characterize different failure modes: agreement alone treats strongly and weakly supported answers identically, whereas likelihood alone may remain high even when the re-elicited answer differs from the primary answer.

\subsection{\tup (\tupm)}\label{sec:tup}

\subsubsection{Token-Level Uncertainty Signals}
To complement \pcem, we design \tupm as a fixed-length, position-aware representation of the uncertainty trajectory in a variable-length response. It summarizes token-level signals already available from the completed response and requires no additional decoding. Let $y=(y_1,\ldots,y_T)$ denote the generated response. For each token $y_t$, we retain its selected-token log-probability $\lambda_t$ and renormalized top-$k$ entropy $\eta_t$:
\begin{equation}
\lambda_t=\log p_\theta(y_t\mid x,y_{<t}),\qquad
\eta_t=-\sum_{v\in\mathcal{V}^{(k)}_t}\widetilde p_t(v)\log\widetilde p_t(v).
\label{eq:tup_token}
\end{equation}
Here, $\widetilde p_t$ is the predictive distribution renormalized over the top-$k$ support $\mathcal{V}^{(k)}_t$.

\subsubsection{Position-Aware Uncertainty Profile}
To preserve coarse trajectory shape across responses of different lengths, let $J$ denote the number of normalized-position bins. For each sequence $z\in\{\lambda,\eta\}$, we define
\begin{equation}
\begin{aligned}
b_j(z)&=\frac{1}{|\mathcal B_j|}\sum_{t\in\mathcal B_j}z_t,\\
\mathcal B_j&=\left\{t:\frac{j-1}{J}<\frac{t}{T}\leq\frac{j}{J}\right\},
\qquad j=1,\ldots,J.
\end{aligned}
\label{eq:tup_bins}
\end{equation}
The vector $\mathbf{b}(z)=[b_1(z),\ldots,b_J(z)]\in\mathbb{R}^{J}$ maps an arbitrary-length sequence to a resolution-controlled profile. The resulting $2J$ bin summaries capture the position-dependent shapes of both log-probability and entropy from the beginning of the reasoning trace to the final answer.

\subsubsection{Global and Trace Statistics}
Beyond these position-wise features, \tupm records complementary global, trend, and boundary-sensitive summaries. We define
\begin{equation}
\begin{aligned}
\mathbf{g}(\lambda)
&=[\mu_\lambda,\min_t\lambda_t,\sigma_\lambda,
\beta_\lambda,R_\lambda^2],\\
\mathbf{g}(\eta)
&=[\mu_\eta,\max_t\eta_t,\beta_\eta,R_\eta^2],
\end{aligned}
\label{eq:tup_global}
\end{equation}
where $\mu_z$ denotes the sequence mean, $\sigma_\lambda$ denotes the standard deviation of the log-probability sequence, and $\beta_z$ and $R_z^2$ are the slope and goodness of fit obtained by regressing $z_t$ on normalized token position $t/T$. The order of the log-probability statistics follows the implementation: mean, minimum, and standard deviation precede the trend statistics.

Let $\mu_{\mathrm{tail}}(W)$ be the mean selected-token log-probability over the final $W$ tokens, and let $\rho_{\mathrm{low}}(\tau)$ be the fraction of tokens with $\lambda_t<\tau$. We collect the non-binned summaries as
\begin{equation}
\mathbf{s}=
[\mathbf{g}(\lambda);\mathbf{g}(\eta);
\mu_{\mathrm{tail}}(W);\rho_{\mathrm{low}}(\tau)]
\in\mathbb{R}^{d_{\mathrm{aux}}},
\label{eq:tup_auxiliary}
\end{equation}
where $d_{\mathrm{aux}}$ denotes their total dimension. The complete passive representation is
\begin{equation}
\phi_P=[\mathbf{b}(\lambda);\mathbf{b}(\eta);\mathbf{s}]
\in\mathbb{R}^{d_P},
\qquad
d_P=2J+d_{\mathrm{aux}}.
\label{eq:tup_profile}
\end{equation}
Thus, \tupm separates a resolution-controlled description of trajectory shape from global and boundary-sensitive evidence while producing a fixed-dimensional representation. The values of $J$, $k$, $W$, $\tau$, $d_{\mathrm{aux}}$, and $d_P$ are fixed across all examples and reported in Appendix~\ref{app:setup}.

\subsection{Unified Scoring and Calibration}

\subsubsection{Representation Fusion}
We concatenate the active and passive representations to obtain the unified \m score:
\begin{equation}
u_{\m}=\sigma\!\left(\mathbf w^\top[\phi_A;\phi_P]+b\right).
\label{eq:trac}
\end{equation}
\tupm characterizes uncertainty in the generated response, whereas \pcem actively probes whether the model continues to support its returned answer.
The two modules also differ operationally. \tupm is computed directly from the token scores returned during the primary generation. \pcem performs one short cached decode and stores an explicit record containing the re-elicited answer, its agreement with the primary answer, and its likelihood and confidence statistics. This auditable interface allows answer normalization, parsing rules, and feature definitions to be revised without regenerating the primary reasoning trace. 

\subsubsection{Correctness-Supervised Calibration}
Given a calibration set $\mathcal D=\{(\phi_i,z_i)\}_{i=1}^N$, each learned variant minimizes the same regularized binary cross-entropy objective:
\begin{equation}
\min_{\mathbf w,b}\frac{1}{N}\sum_i
-z_i\log u_i-(1-z_i)\log(1-u_i)
+\lambda\lVert\mathbf w\rVert_2^2.
\label{eq:training}
\end{equation}

Features are standardized using only the training fold. The model remains frozen, and online scoring uses no verifier, reference answer, process label, hidden-state encoder, or LLM judge.
Logistic regression keeps the estimator lightweight and enables interpretable comparisons among matched representations. Restricting auxiliary model capacity also helps attribute performance gains to the uncertainty observations rather than to a more expressive classifier. 

\subsubsection{Offline Preparation and Training}

Primary responses are generated once, with the required token log-probabilities and entropy statistics retained. \pcem stores an immutable record of the re-elicited answer and its probability statistics rather than only the resulting feature vector. Answer normalization, equivalence rules, and representation choices can therefore be revised without regenerating the expensive primary response. Deterministic verification is applied only after the primary answer has been stored and provides the binary label used for calibration.
Within each training fold, missing answer fields are assigned an explicit missingness indicator, continuous features are standardized, and the regularized logistic head is fitted to the resulting representation. The missing-value convention, standardizer, and head parameters are saved together. No test-fold statistic enters preprocessing. This fold-local procedure is shared by the active, passive, unified, and fusion variants, ensuring that differences among them reflect their uncertainty observations rather than different fitting pipelines.

\subsubsection{Online Inference}
At deployment time, the frozen model first produces the primary reasoning trace, final answer, and token scores. \tupm is computed directly from these scores. \pcem then reuses the cached prefix to decode a short answer suffix, after which the primary and re-elicited answers are mapped to deterministic equivalence classes before $\phi_A$ is constructed. The saved preprocessing pipeline and prediction head produce $u_{\m}$ without access to the reference answer $a^*$ or verifier $V$. The resulting score can support abstention, prioritize responses for human review, or allocate additional samples to low-scoring responses.



\subsection{\cf (\cfm)}

When $K$ complete reasoning traces are available, we define the answer frequency and conventional self-consistency confidence as
\begin{equation}
\pi_b=\frac{1}{K}\sum_{k=1}^{K}
\mathbb{I}[a^{(k)}\equiv b],
\qquad
q_K=\max_b\pi_b,
\label{eq:sc}
\end{equation}
where $q_K$ is the modal-answer vote fraction used by standard $K$-sample self-consistency (SC@$K$). Because our task evaluates the correctness of the primary answer $a$, we additionally report the response-aligned baseline
\begin{equation}
q_K(a)=\frac{1}{K}\sum_{k=1}^{K}
\mathbb{I}[a^{(k)}\equiv a].
\label{eq:primary_sc}
\end{equation}

Beyond these baselines, \cfm combines the within-trace representations of \m with cross-trace consensus. We pair $q_K$ with the answer entropy
$h_K=-\sum_b\pi_b\log\pi_b$ and compute
\begin{equation}
u_{\cfm}=\sigma\!\left(
\mathbf w_f^\top
[\phi_A;\phi_P;q_K;h_K]+b_f
\right).
\label{eq:fusion}
\end{equation}
Because the primary trace is one of the $K$ samples, \cfm uses the same $K$ complete generations as SC@$K$, together with one short cached probe. It tests whether within-trace re-elicitation adds information beyond cross-trace consensus. Unlike the default \m score, which targets low-latency estimation from one complete trace, \cfm augments an ensemble that is already available.

\section{Evaluation}\label{sec:evaluation}
We organize the experiments around five research questions (RQs):
\begin{itemize}
    \item \textbf{RQ 1:} How does \m compare with single-trace and sampling-based baselines in correctness ranking and selective risk?
    \item \textbf{RQ 2:} What does each view contribute, does re-elicitation depend on the completed trace, and is the resulting signal robust to decoding choices?
    \item \textbf{RQ 3:} Can re-elicitation help identify errors when sampled answers reach consensus?
    \item \textbf{RQ 4:} How robust, transferable, and label-efficient is the re-elicitation signal?
    \item \textbf{RQ 5:} What computational overhead does cached re-elicitation introduce?
\end{itemize}

\subsection{Experimental Setup}
\subsubsection{Datasets}
We evaluate \m on five mathematical reasoning benchmarks: GSM8K~\cite{cobbe2021gsm8k}, MATH500~\cite{hendrycks2021math}, Minerva~\cite{lewkowycz2022minerva}, OlympiadBench~\cite{he2024olympiad}, and AIME~\cite{aime24}. To assess its generalizability beyond mathematical reasoning, we additionally evaluate \m on BIG-Bench Hard~\cite{suzgun2023bbh} and GPQA-Diamond~\cite{rein2024gpqa}.

\subsubsection{Backbone LLMs}
We evaluate six reasoning-model configurations across three LLM families: Qwen3-4B~\cite{yang2025qwen3}, Qwen3-8B~\cite{yang2025qwen3}, Qwen3-14B~\cite{yang2025qwen3}, Qwen3.5-9B~\cite{qwen2026qwen35}, Phi-4-reasoning~\cite{abdin2024phi4}, and Ministral-3-14B-Reasoning~\cite{mistral2026ministral3}.

\subsubsection{Baselines}
Probability-based single-trace baselines include mean token log-probability, Self-certainty~\cite{kang2025selfcertainty}, and DeepConf-bottom~\cite{fu2025deepconf}. We also compare P(True)~\cite{kadavath2022mostly} and SelfDoubt~\cite{pandey2026selfdoubt}, which use elicited self-verification and verbalized uncertainty, respectively. Component baselines include \tupm, the supervised passive trace-profile component, and \pcem, the active answer re-elicitation component. Sampling baselines include SelfCheckGPT@8~\cite{manakul2023selfcheckgpt} and eight-sample self-consistency (SC@8)~\cite{wang2023selfconsistency}, using agreement across independently sampled responses. 
Supervised Consensus@8 is a supervised baseline that pools vote fraction, answer entropy, vote margin, trace log-probability, passive \tupm features, and answer likelihood using the same logistic head and correctness supervision as \m. All learned methods share the same folds and preprocessing, and results are averaged equally across model--dataset pairs.

\subsubsection{Metrics} \label{sec:metrics}
We evaluate uncertainty estimation using two complementary metrics.
AUROC measures how effectively a score ranks correct responses above
incorrect ones, with higher values indicating better discrimination.
Let $(x^+,y^+)$ and $(x^-,y^-)$ denote randomly selected correct and
incorrect responses, respectively. Then
\begin{equation}
\begin{aligned}
\operatorname{AUROC}(u)
={}&
\Pr\bigl[u(x^+,y^+)>u(x^-,y^-)\bigr] \\
&+
\tfrac{1}{2}
\Pr\bigl[u(x^+,y^+)=u(x^-,y^-)\bigr].
\end{aligned}
\label{eq:auroc}
\end{equation}

To evaluate selective prediction, let $N$ responses be ordered by
decreasing score,
$u_{\pi(1)}\geq\cdots\geq u_{\pi(N)}$.
At coverage $c_k=k/N$, the selective risk and area under the
risk--coverage curve (AURC) are
\begin{equation}
R(c_k)
=
\frac{1}{k}
\sum_{j=1}^{k}
\bigl(1-z_{\pi(j)}\bigr),
\qquad
\operatorname{AURC}(u)
=
\frac{1}{N}
\sum_{k=1}^{N}R(c_k).
\label{eq:aurc}
\end{equation}

Lower AURC indicates that incorrect responses receive lower scores and
are rejected earlier as coverage decreases. AUROC evaluates pairwise
correctness ranking, whereas AURC measures the error retained across
different coverage levels. 

\subsubsection{Protocol}
Deterministic verification supplies offline labels. Learned estimators use five shuffled five-fold splits with question-level separation, and every in-domain score is out-of-fold. All variants share the logistic head, folds, standardization, cached generations, and labels. Headline intervals use a pair-equal hierarchical bootstrap with 2,000 resamples. Appendix~\ref{app:setup} shows complete model, data, baseline, and fitting details.

\begin{table*}[t]
\centering
\small
\caption{
Overall uncertainty-estimation performance. Macro reports pair-equal AUROC/AURC; excess AURC normalizes selective risk by the base error rate, and risk at coverage is the error rate over the most-confident responses. SC denotes self-consistency. Best results are \textbf{bold} with second-best results \underline{underlined}. Relative latency is measured on Qwen3-8B (Section~\ref{sec:rq5}).}
\setlength{\tabcolsep}{3pt}

\def\apair#1#2{#1\ /\ #2}

\begin{tabular}{llc ccccc cc ccc}
\toprule
& \multirow{2}{*}{\textbf{Method}}
& \multirow{2}{*}{\textbf{Latency}}
& \multicolumn{6}{c}{\textbf{AUROC$\uparrow$/AURC$\downarrow$}}
& \multicolumn{1}{c}{\textbf{AURC}}
& \multicolumn{3}{c}{\textbf{Risk at coverage} $\downarrow$}
\\
\cmidrule(lr){4-9}
\cmidrule(lr){10-10}
\cmidrule(lr){11-13}
&
&
& GSM8K
& MATH500
& Minerva
& Olympiad
& AIME
& Macro
& Excess
& 10\%
& 20\%
& 50\%
\\
\midrule

\multirow{7}{*}{\rotatebox{90}{\textit{Single}}}
& Mean log-probability
& $1.00\times$
& \apair{.822}{.024}
& \apair{.627}{.190}
& \apair{.561}{.517}
& \apair{.688}{.322}
& \apair{.663}{.555}
& \apair{.672}{.322}
& .696
& .272
& .296
& .324
\\

& Self-certainty
& $1.00\times$
& \apair{.826}{.024}
& \apair{.632}{.192}
& \apair{.561}{.519}
& \apair{.700}{.318}
& \apair{.671}{.556}
& \apair{.678}{.322}
& .692
& .264
& .296
& .321
\\

& DeepConf-bottom
& $1.00\times$
& \apair{.648}{.041}
& \apair{.521}{.238}
& \apair{.530}{.568}
& \apair{.624}{.370}
& \apair{.625}{.597}
& \apair{.590}{.363}
& .875
& .365
& .343
& .350
\\

& P(True)
& $+1$ fwd
& \apair{.773}{.022}
& \apair{.822}{.099}
& \apair{\underline{.720}}{\underline{.410}}
& \apair{.769}{.259}
& \apair{.831}{.423}
& \apair{.783}{.243}
& .530
& \underline{.106}
& .143
& .247
\\

& SelfDoubt
& $1.00\times$
& \apair{.842}{.019}
& \apair{.804}{.112}
& \apair{.692}{.433}
& \apair{.827}{.224}
& \apair{.714}{.532}
& \apair{.776}{.264}
& .538
& .161
& .184
& .262
\\

& \tupm\ (passive)
& $1.00\times$
& \apair{\underline{.855}}{\underline{.017}}
& \apair{.819}{.106}
& \apair{.704}{.422}
& \apair{.843}{.212}
& \apair{.726}{.520}
& \apair{.789}{.255}
& \underline{.514}
& .153
& .175
& .253
\\

& \pcem\ (active)
& $1.02\times$
& \apair{.741}{.033}
& \apair{\underline{.901}}{\underline{.076}}
& \apair{.695}{.447}
& \apair{\underline{.905}}{\underline{.191}}
& \apair{\underline{.926}}{\underline{.402}}
& \apair{\underline{.834}}{\underline{.230}}
& .551
& .129
& \underline{.135}
& \underline{.207}
\\

\rowcolor{gray!15}
\cellcolor{white}
& \textbf{\m}
& $1.02\times$
& \apair{\textbf{.896}}{\textbf{.011}}
& \apair{\textbf{.931}}{\textbf{.064}}
& \apair{\textbf{.760}}{\textbf{.379}}
& \apair{\textbf{.940}}{\textbf{.164}}
& \apair{\textbf{.944}}{\textbf{.370}}
& \apair{\textbf{.894}}{\textbf{.198}}
& \textbf{.392}
& \textbf{.055}
& \textbf{.093}
& \textbf{.184}
\\

\midrule

\multirow{3}{*}{\rotatebox{90}{\textit{8-samp.}}}

& SelfCheckGPT@8
& $2.23\times$
& \apair{.836}{.024}
& \apair{.919}{.076}
& \apair{.748}{.414}
& \apair{.895}{.190}
& \apair{.907}{.379}
& \apair{.861}{.217}
& .451
& .091
& .116
& .211
\\

& Supervised Consensus@8
& $2.16\times$
& \apair{.845}{.021}
& \apair{.930}{.071}
& \apair{.758}{.402}
& \apair{.906}{.182}
& \apair{.918}{.369}
& \apair{.871}{.209}
& .437
& .079
& .105
& .198
\\

& SC@8 (primary answer)
& $2.39\times$
& \apair{\underline{.856}}{.023}
& \apair{.923}{.071}
& \apair{.763}{.410}
& \apair{.899}{.184}
& \apair{.900}{.367}
& \apair{.867}{.216}
& .439
& .085
& .110
& .207
\\

& SC@8 (modal-answer)
& $2.39\times$
& \apair{.849}{\underline{.020}}
& \apair{\underline{.937}}{\underline{.068}}
& \apair{\underline{.764}}{\underline{.395}}
& \apair{\underline{.914}}{\underline{.178}}
& \apair{\underline{.925}}{\underline{.363}}
& \apair{\underline{.878}}{\underline{.205}}
& \underline{.428}
& \underline{.072}
& \underline{.099}
& \underline{.192}
\\

\rowcolor{gray!15}
\cellcolor{white}
& \textbf{\m\ (+\cfm@8)}
& $2.41\times$
& \apair{\textbf{.902}}{\textbf{.011}}
& \apair{\textbf{.962}}{\textbf{.058}}
& \apair{\textbf{.794}}{\textbf{.357}}
& \apair{\textbf{.954}}{\textbf{.158}}
& \apair{\textbf{.965}}{\textbf{.356}}
& \apair{\textbf{.916}}{\textbf{.188}}
& \textbf{.369}
& \textbf{.045}
& \textbf{.067}
& \textbf{.179}
\\

\bottomrule
\end{tabular}

\label{tab:main}
\end{table*}

\subsection{RQ 1: Overall Performance}\label{sec:rq1}
\textbf{\m outperforms eight-sample self-consistency at nearly the latency of one generation.} Table~\ref{tab:main} reports dataset-level AUROC for single-trace and eight-sample methods. \m improves on every single-trace baseline; active \pcem alone ranks above all prior single-trace estimators.
SC@8 observes eight completed trajectories, whereas \m observes one trajectory plus one short cached answer, yet the paired difference favors \m ($+0.016$, 95\% CI $[+0.010,+0.022]$, winning on most pairs). Thus, \m outperforms SC@8 while decoding one rather than eight full trajectories. The advantage concentrates where consensus is least reliable, reaching $+0.047$ on GSM8K, while MATH500 and Minerva slightly favor SC@8. Table~\ref{tab:main} also reports the primary-answer vote share $q_K(a)$, ruling out an artifact from scoring SC@8 by its modal answer rather than the primary answer. This variant is weaker than modal consensus (macro AUROC $0.867$ versus $0.878$), so we use the stronger modal form, which \m also leads.

The ranking advantage also yields lower selective risk. \m gives the lowest single-trace values. At 10\% coverage, it reduces error by about 24\% relative to SC@8 at $8\times$ the cost. Figure~\ref{fig:riskcov} shows that \m tracks SC@8 across coverage levels while both outperform P(True) and SC. Figure~\ref{fig:pareto} shows that \m reaches SC@8 quality at $1.02\times$ cost, while fusion extends the frontier at a matched eight-sample budget. \textbf{Answer to RQ 1:} \m is the strongest single-trace estimator that outperforms eight-sample consensus at 1.02$\times$ latency with the lowest single-trace selective risk.

\begin{table}[t]
\centering
\small
\caption{
Macro ablation results. The 95\% confidence intervals are
computed using a bootstrap with 2,000 resamples.
}
\label{tab:ablation}
\setlength{\tabcolsep}{4pt}
\begin{tabular}{lccc}
\toprule
\textbf{Representation}
& \textbf{AUROC $\uparrow$}
& \textbf{CI of AUROC}
& \textbf{AURC $\downarrow$} \\
\midrule

\multicolumn{4}{l}{\textbf{(a) Component contribution}} \\
\tupm\ only
& .789 & [.771, .807] & .255 \\
\pcem\ only
& .834 & [.816, .851] & .230 \\
\rowcolor{gray!15}
\pcem $+$ \tupm\ ($=$ \m)
& \textbf{.894} & \textbf{[.880, .907]} & \textbf{.198} \\

\midrule
\multicolumn{4}{l}{\textbf{(b) Trace-conditioning controls}} \\
Question-only probe
& .734 & [.711, .756] & .292 \\
Shuffled-trace probe
& .681 & [.655, .706] & .319 \\
Teacher-forced support
& .756 & [.734, .777] & .258 \\
PCE $+$ answer support
& .813 & [.793, .832] & .241 \\
\pcem\ with completed trace
& \textbf{.834} & \textbf{[.816, .851]} & \textbf{.230} \\

\midrule
\multicolumn{4}{l}{\textbf{(c) PCE feature decomposition}} \\
Agreement only
& .751 & [.729, .772] & .280 \\
Likelihood only
& .726 & [.703, .748] & .298 \\
Agreement $+$ likelihood (\pcem)
& \textbf{.834} & \textbf{[.816, .851]} & \textbf{.230} \\

\midrule
\multicolumn{4}{l}{\textbf{(d) Comparison with self-verification}} \\
P(True)
& .783 & [.762, .803] & .243 \\
\pcem
& .834 & [.816, .851] & .230 \\
P(True) $+$ \pcem
& \textbf{.859} & \textbf{[.842, .875]} & \textbf{.216} \\

\bottomrule
\end{tabular}
\end{table}
\begin{table}[t]
\centering
\small
\caption{Decoding-symmetry matrix on Qwen3-8B, varying primary-generation and probe temperature. Agreement is the primary-to-re-elicited match rate.
}
\setlength{\tabcolsep}{9pt}
\begin{tabular}{lrrrr}
\toprule
\textbf{Primary $\to$ probe} & \textbf{Acc} & \textbf{Agr.} & \textbf{\pcem} & \textbf{\m} \\
\midrule
\multicolumn{5}{l}{\textbf{MATH500 (Qwen3-8B, $n{\approx}295$)}}\\
$T{=}0.7\to$ greedy (default) & .719 & .753 & \underline{.780} & \textbf{.803} \\
greedy $\to$ greedy & .707 & .771 & \underline{.773} & \textbf{.829} \\
$T{=}0.7\to T{=}0.7$ & .719 & .759 & \textbf{.812} & \underline{.809} \\
$T{=}0.2\to$ greedy & .724 & .761 & \underline{.798} & \textbf{.820} \\
$T{=}1.0\to$ greedy & .711 & .759 & \underline{.764} & \textbf{.855} \\
\midrule
\multicolumn{5}{l}{\textbf{Minerva (Qwen3-8B, $n{\approx}260$)}}\\
$T{=}0.7\to$ greedy (default) & .307 & .674 & \underline{.801} & \textbf{.804} \\
greedy $\to$ greedy & .296 & .654 & \underline{.793} & \textbf{.828} \\
$T{=}0.7\to T{=}0.7$ & .307 & .674 & \underline{.801} & \textbf{.802} \\
$T{=}0.2\to$ greedy & .290 & .634 & \underline{.774} & \textbf{.814} \\
$T{=}1.0\to$ greedy & .324 & .652 & \textbf{.758} & \underline{.751} \\
\bottomrule
\end{tabular}
\label{tab:decmatrix}
\end{table}

\subsection{RQ 2: Ablation Study}\label{sec:rq2}

\textbf{(a) Active and passive views are complementary.}
Combining \tupm\ and \pcem\ substantially outperforms either component and achieves the best overall performance. This confirms that token-level uncertainty dynamics and trace-conditioned answer support capture complementary rather than redundant evidence.
\textbf{(b) Effective re-elicitation requires the corresponding trace.}
Question-only, shuffled-trace, and teacher-forced variants all underperform \pcem. The gain therefore comes specifically from re-eliciting an answer under its corresponding completed reasoning trace, rather than independently resolving the question, introducing arbitrary reasoning context, or merely rescoring the original answer.
\textbf{(c) Agreement and probabilistic support are both useful.}
Agreement indicates whether re-elicitation changes the original conclusion, while likelihood and confidence measure how strongly the re-elicited answer is supported. Their combination clearly outperforms either signal alone, showing that answer identity and support strength capture different aspects of uncertainty.
\textbf{(d) Re-elicitation differs from explicit self-verification.}
\pcem\ outperforms P(True), while combining the two yields further improvements. This indicates that trace-conditioned re-elicitation and explicit correctness judgments capture related but non-redundant evidence about response reliability.
\textbf{(Table~\ref{tab:decmatrix}) The signal is robust to decoding choices.}
Both \pcem\ and the full \m\ remain informative under fully greedy decoding and perform broadly consistently across alternative temperature settings. Agreement also remains below saturation, indicating that the signal does not depend solely on sampling variation. Instead, it reflects trace-conditioned answer support rather than an artifact of a particular decoding configuration.

\textbf{Answer to RQ 2:}
Active and passive views provide complementary uncertainty evidence. Effective re-elicitation depends on the corresponding completed trace, combines answer agreement with probabilistic support, differs from explicit self-verification, and remains informative under deterministic decoding.

\subsection{RQ 3: Consensus Coverage}\label{sec:rq3}
\textbf{Active re-elicitation remains informative when vote confidence has no resolution and improves the matched eight-generation system.} 
The bottom block of Table~\ref{tab:main} tests the practical implication. \m (+\cfm) uses the same eight full generations plus one short cached probe and raises macro AUROC by $+0.038$ (95\% CI $[+0.022,+0.055]$, $p<0.001$), with gains on most pairs, while also lowering macro AURC. 
This matched-budget comparison shows that re-elicitation supplies information unavailable to identical-budget consensus. Thus, the gain comes from the active observation rather than merely replacing a raw scalar with supervised passive consensus features. 
As shown in Table~\ref{tab:app-blindspot}, 
at an SC@8 vote threshold of one, all sampled answers are identical, so the vote fraction is constant, and its AUROC is 0.500 by construction. 
Among all pairs containing unanimous cases, 7,485 responses are unanimous, and 243 are still wrong. \pcem remains discriminative because re-elicitation and its confidence can vary even when all completed answers agree.
Appendix~\ref{app:fusion} reports thresholds, per-pair results, and alternative fusion designs. 
\textbf{Answer to RQ 3:} re-elicitation provides ranking information after the vote fraction saturates and improves the matched eight-generation performance.

\begin{table}[H]
\centering
\caption{Cross-dataset and cross-model transfer. Top: per-target leave-one-dataset-out (LODO) for the \pcem head, with the five-fold in-domain score for reference. Bottom: the three representations under LODO, a single global head (leave-one-pair-out), and leave-one-model-out (LOMO).}
\label{tab:transfer}
\small
\setlength{\tabcolsep}{12pt}
\begin{tabular}{lrrr}
\toprule
\textbf{Held-out target} & \textbf{Transfer} & \textbf{In-domain} & \textbf{$\Delta$} \\
\midrule
GSM8K & \textbf{.772} & .741 & $+.031$ \\
MATH500 & \textbf{.909} & .901 & $+.007$ \\
Minerva & .669 & \textbf{.695} & $-.026$ \\
OlympiadBench & \textbf{.914} & .905 & $+.009$ \\
AIME & \textbf{.966} & .926 & $+.040$ \\
\midrule
\multicolumn{4}{l}{\textbf{Representation comparison (macro)}} \\
& LODO & Global & LOMO \\
\quad \tupm & .799 & .774 & .759 \\
\quad \pcem & .846 & .827 & .826 \\
\rowcolor{gray!15}
\quad \m & \textbf{.897} & \textbf{.893} & \textbf{.884} \\
\bottomrule
\end{tabular} 
\end{table}

\subsection{RQ 4: Generality}\label{sec:rq4}

\textbf{The active signal transfers across mathematical datasets and remains stable under acquisition changes.} Three generation seeds, trace-length matching, and semantically equivalent probe cues preserve its advantage. A leave-one-dataset-out head matches or exceeds its in-domain counterpart on the active view for four of the five held-out targets as shown in Table~\ref{tab:transfer}, while a single global head without per-pair fitting remains within noise; transfer to held-out AIME is strongest and exceeds its in-domain score. The full \m estimator also leads \pcem and \tupm under leave-one-dataset-out, global-head, and leave-one-model-out evaluation, with little loss under cross-model reuse. Thus, the signal is not tied to per-pair feature engineering.
The robustness tests vary both the generated trace and the measurement procedure. Across three generation seeds, the active advantage remains positive. Three equivalent answer cues yield a small AUROC spread, and length-matched analysis preserves the trend, indicating that the head does not depend on exact cue wording or simply learn that longer traces are less reliable. We restrict headline claims to verifiable mathematical reasoning; transfer across model families and to non-mathematical domains is mixed and reported in Appendix~\ref{app:robustness}.

\begin{table}[H]
\centering
\small
\caption{Label efficiency on a fixed 40\% test split.}
\setlength{\tabcolsep}{12pt}
\begin{tabular}{rrrr}
\toprule
\textbf{Labels} & \textbf{Sequence only} & \textbf{Conv.+final} & \textbf{\pcem} \\
\midrule
1\% & .651 & \underline{.691} & \textbf{.736} \\
5\% & .652 & \underline{.694} & \textbf{.741} \\
25\% & .669 & \underline{.711} & \textbf{.763} \\
100\% & .674 & \underline{.737} & \textbf{.796} \\
\bottomrule
\end{tabular}
\label{tab:label-efficiency}
\end{table}

The active observation is also label-efficient. Table~\ref{tab:label-efficiency} varies the labeled fraction on a fixed 40\% test split: with 1\% of labels, the active view matches a re-answer endpoint control trained with all labels, and at 25\% it retains 95.9\% of its full-data score. Because only the calibration sample changes, the fixed representation provides a favorable starting geometry for a low-capacity head rather than relying on large supervised sets, benefiting deployments that can label a modest calibration set but cannot retrain the reasoning model. \textbf{Answer to RQ 4:} re-elicitation transfers across mathematical datasets with negligible degradation and retains most of its quality with a fraction of the calibration labels, while new model families or domains benefit from representative data.

\begin{figure}[t]
    \centering
    \subfigure[Macro risk-coverage (lower is better).\label{fig:riskcov}]{
      \includegraphics[width=0.47\linewidth]{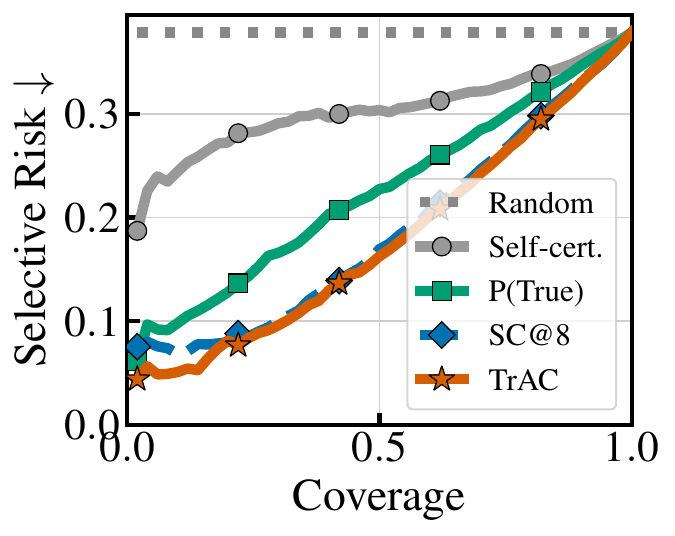}}
    \hfill
    \subfigure[Compute-quality tradeoff.\label{fig:pareto}]{%
      \includegraphics[width=0.47\linewidth]{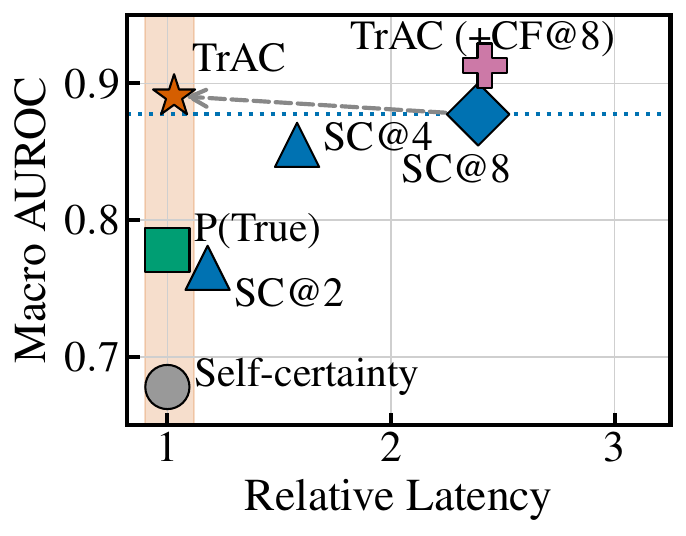}}
    \caption{Effectiveness and cost of \m. (a) \m tracks SC@8 across coverage levels. (b) \m reaches SC@8 quality at $1.02\times$ latency and dominates lower-budget sampling, while fusion extends the frontier at a matched budget.}
    \Description{}
    \label{fig:analysis} 
\end{figure}

\subsection{RQ 5: Efficiency}\label{sec:rq5}
\begin{table}[t]
\centering
\small
\caption{Measured end-to-end efficiency on Qwen3-8B.}
\setlength{\tabcolsep}{6pt}
\begin{tabular}{lccrr}
\toprule
& \multicolumn{2}{c}{\textbf{Median latency (s)}} & \textbf{Rel.} & \textbf{Macro} \\
\cmidrule(lr){2-3}
\textbf{Estimator} & MATH500 & Olympiad & \textbf{lat.} & \textbf{AUROC} \\
\midrule
Self-certainty & 1.205 & 2.782 & $1.00\times$ & .678 \\
P(True) & 1.208 & 2.789 & $1.00\times$ & .778 \\
\rowcolor{gray!15}
\textbf{\m} & 1.239 & 2.876 & $1.02\times$ & \textbf{.894} \\
\rowcolor{gray!15}
 \textbf{\m, cache off} & 1.370 & 3.302 & $1.16\times$ & \textbf{.894} \\
SC@2 & 1.300 & 3.574 & $1.18\times$ & .765 \\
SC@4 & 1.565 & 5.164 & $1.58\times$ & .855 \\
SC@8 & 2.091 & 8.444 & $2.39\times$ & .878 \\
\rowcolor{gray!15}
\textbf{\m(+\cfm)} & 2.102 & 8.394 & $2.41\times$ & \textbf{.916} \\
\bottomrule
\end{tabular}
\label{tab:pareto-main}
\end{table}

\textbf{Cached re-elicitation adds modest measured latency and lies on the observed accuracy-latency frontier.} As shown in Table~\ref{tab:pareto-main}, re-elicitation requires $1.02\times$ the primary-generation latency on Qwen3-8B, versus $2.39\times$ for SC@8 averaged over MATH500 and Olympiad serving profiles. Passive \tupm requires no additional model call. SC@2 is dominated by \m, while SC@4, SC@8, and \cfm trade progressively more latency for higher AUROC, consistent with Figure~\ref{fig:pareto}.
Cache reuse is central to this operating point: without it, the completed prefix must be reprocessed, whereas the cached implementation reuses prefix states and decodes only a short answer suffix. Appendix~\ref{app:efficiency} reports raw timings, throughput, hardware, and the cache-off comparison. \textbf{Answer to RQ 5:} \m adds 2\% measured latency in the audited cached system and gives the best single-trace results at that cost.
Together, the operating modes support a direct compute policy without changing the underlying estimator. With one trace, \m extracts active and passive evidence at near-single-generation latency and outperforms eight-sample consensus. When eight traces are already sampled, \cfm adds the active observation and targets unanimous errors. A deployment can therefore choose between two budgets without changing the re-elicitation representation.
Token counts and latency are reported separately because scheduling, batching, and cache management prevent direct conversion. More details are in Appendix~\ref{app:efficiency}. 

\section{Related Work}
\label{sec:related-work}

\subsection{Passive Single-Trace UQ}

Single-trace uncertainty estimators for LLMs typically rely on token probabilities, verbalized confidence, or learned representations. Early question-answering studies calibrate answer likelihoods or train models to express uncertainty~\cite{jiang2021know,lin2022verbalized,azaria2023internal}, while later work systematically compares confidence-elicitation strategies~\cite{xiong2024express}. P(True) asks the model whether a proposed answer is correct~\cite{kadavath2022mostly}; self-certainty measures concentration in the predictive distribution; and DeepConf aggregates evidence from low-confidence token windows~\cite{kang2025selfcertainty,fu2025deepconf}. LM-Polygraph benchmarks probability- and representation-based estimators under standardized protocols~\cite{vashurin2025polygraph}, and recent studies evaluate confidence estimation for large reasoning models in high-stakes domains~\cite{khanmohammadi2026reliable}. Other work characterizes how token-level uncertainty evolves throughout reasoning traces~\cite{gruenefeld2026tracing,ballon2026probing,wang2026cdg}. Most of these methods passively score token probabilities, generated text, or hidden representations already produced during the original generation~\cite{duan2024shifting,su2024unsupervised}. In contrast, \m combines active answer re-elicitation through \pcem with the passive token-level profile of \tupm to capture complementary uncertainty evidence from one completed reasoning trace.

\subsection{Vote-Based Sampling UQ}

Sampling-based uncertainty estimators compare multiple independently generated responses~\cite{lin2023generating}. Self-consistency measures confidence using the vote share of the most frequent final answer across sampled reasoning traces~\cite{wang2023selfconsistency}. SelfCheckGPT similarly uses cross-sample consistency for black-box hallucination detection~\cite{manakul2023selfcheckgpt}, while semantic entropy groups meaning-equivalent outputs before quantifying disagreement~\cite{kuhn2023semantic}. Confidence-informed variants weight or filter sampled traces according to their estimated reliability~\cite{taubenfeld2025cisc}. These methods capture variation across reasoning traces, but require multiple complete generations and lose ranking resolution when sampled answers reach consensus; in particular, a unanimous vote assigns the same confidence to a robustly correct answer and an error repeatedly produced across samples. In contrast, \pcem measures within-trace answer support through one short re-elicitation and remains informative when cross-trace consensus saturates.

\subsection{Active Prefix Diagnostics}

Answer-convergence methods close partial reasoning chains and track when their predicted answers stabilize, primarily for early stopping~\cite{liu2025convergence}. Confidence Leaps truncates reasoning at intermediate points, elicits an answer, and estimates a distribution over a finite candidate set~\cite{tikhonov2026confidence}. Finite-answer preference stabilization similarly examines when predefined answer verbalizers become stable~\cite{zhang2026commit}. These studies demonstrate that answers can be elicited from partial reasoning prefixes, but primarily focus on identifying stopping points or preference transitions. In contrast, our method re-elicits an open-form answer from the completed reasoning prefix to estimate the correctness of the returned response through answer consistency and probabilistic support.
\section{Conclusion}\label{sec:conclusion}

In this paper, we introduce \m, a correctness-supervised UQ framework based on \emph{trace-conditioned answer consistency}. \m combines two complementary views derived from one completed reasoning trace. Rather than generating another complete reasoning trace or invoking a separate judge, \pce actively re-elicits a short answer from the completed reasoning prefix and represents both its consistency with the original answer and its probabilistic support. This active representation is combined with \tup, which summarizes passive token-level uncertainty already available from the original generation. When multiple completed samples are already available, the optional consensus fusion is used to provide additional vote statistics.
Across five mathematical reasoning benchmarks and three LLM families, \m outperforms eight-sample self-consistency using only one complete reasoning trace and a short cached answer probe. It improves macro AUROC by 1.8\% and reduces AURC by 3.4\%, while adding only 2\% measured latency. When eight samples are already available, incorporating re-elicitation into consensus-based estimation further improves macro AUROC by 4.3\% and reduces AURC by 8.3\%, without additional full-trace generation. These results show that trace-conditioned answer consistency is a low-overhead uncertainty signal that complements both passive token statistics and cross-sample consensus, while remaining informative when consensus saturates.

\clearpage
\bibliographystyle{ACM-Reference-Format}
\bibliography{reference}

@inproceedings{wang2023selfconsistency,
  title = {Self-Consistency Improves Chain of Thought Reasoning in Language Models},
  author = {Wang, Xuezhi and Wei, Jason and Schuurmans, Dale and Le, Quoc V. and Chi, Ed H. and Narang, Sharan and Chowdhery, Aakanksha and Zhou, Denny},
  booktitle = {The Eleventh International Conference on Learning Representations},
  year = {2023},
  url = {https://openreview.net/forum?id=1PL1NIMMrw}
}

@inproceedings{manakul2023selfcheckgpt,
  title = {{SelfCheckGPT}: Zero-Resource Black-Box Hallucination Detection for Generative Large Language Models},
  author = {Manakul, Potsawee and Liusie, Adian and Gales, Mark J. F.},
  booktitle = {Proceedings of the 2023 Conference on Empirical Methods in Natural Language Processing},
  pages = {9004--9017},
  year = {2023},
  publisher = {Association for Computational Linguistics},
  doi = {10.18653/v1/2023.emnlp-main.557}
}

@inproceedings{xiong2024express,
  title = {Can {LLM}s Express Their Uncertainty? An Empirical Evaluation of Confidence Elicitation in {LLM}s},
  author = {Xiong, Miao and Hu, Zhiyuan and Lu, Xinyang and Li, Yifei and Fu, Jie and He, Junxian and Hooi, Bryan},
  booktitle = {The Twelfth International Conference on Learning Representations},
  year = {2024},
  url = {https://openreview.net/forum?id=gjeQKFxFpZ}
}

@article{lin2022verbalized,
  title = {Teaching Models to Express Their Uncertainty in Words},
  author = {Lin, Stephanie and Hilton, Jacob and Evans, Owain},
  journal = {Transactions on Machine Learning Research},
  year = {2022},
  url = {https://openreview.net/forum?id=8s8K2UZGTZ}
}

@article{jiang2021know,
  title = {How Can We Know When Language Models Know? On the Calibration of Language Models for Question Answering},
  author = {Jiang, Zhengbao and Araki, Jun and Ding, Haibo and Neubig, Graham},
  journal = {Transactions of the Association for Computational Linguistics},
  volume = {9},
  pages = {962--977},
  year = {2021},
  doi = {10.1162/tacl_a_00407}
}

@misc{kadavath2022mostly,
  title = {Language Models (Mostly) Know What They Know},
  author = {Kadavath, Saurav and Conerly, Tom and Askell, Amanda and Henighan, Tom and Drain, Dawn and Perez, Ethan and Schiefer, Nicholas and Hatfield-Dodds, Zac and DasSarma, Nova and Tran-Johnson, Eli and others},
  note = {arXiv:2207.05221},
  year = {2022},
  doi = {10.48550/arXiv.2207.05221},
  url = {https://arxiv.org/abs/2207.05221}
}

@inproceedings{kang2025selfcertainty,
  title = {Scalable Best-of-{N} Selection for Large Language Models via Self-Certainty},
  author = {Kang, Zhewei and Zhao, Xuandong and Song, Dawn},
  booktitle = {Advances in Neural Information Processing Systems 38},
  year = {2025},
  url = {https://openreview.net/forum?id=nddwJseiiy}
}

@misc{fu2025deepconf,
  title = {Deep Think with Confidence},
  author = {Fu, Yichao and Wang, Xuewei and Tian, Yuandong and Zhao, Jiawei},
  note = {arXiv:2508.15260},
  year = {2025},
  doi = {10.48550/arXiv.2508.15260},
  url = {https://arxiv.org/abs/2508.15260}
}

@inproceedings{taubenfeld2025cisc,
  title = {Confidence Improves Self-Consistency in {LLM}s},
  author = {Taubenfeld, Amir and Sheffer, Tom and Ofek, Eran and Feder, Amir and Goldstein, Ariel and Gekhman, Zorik and Yona, Gal},
  booktitle = {Findings of the Association for Computational Linguistics: ACL 2025},
  pages = {20090--20111},
  year = {2025},
  publisher = {Association for Computational Linguistics},
  doi = {10.18653/v1/2025.findings-acl.1030},
  url = {https://aclanthology.org/2025.findings-acl.1030/}
}

@inproceedings{liu2025convergence,
  title = {Answer Convergence as a Signal for Early Stopping in Reasoning},
  author = {Liu, Xin and Wang, Lu},
  booktitle = {Proceedings of the 2025 Conference on Empirical Methods in Natural Language Processing},
  pages = {17896--17907},
  year = {2025},
  publisher = {Association for Computational Linguistics},
  doi = {10.18653/v1/2025.emnlp-main.904},
  url = {https://aclanthology.org/2025.emnlp-main.904/}
}

@article{vashurin2025polygraph,
  title = {Benchmarking Uncertainty Quantification Methods for Large Language Models with {LM}-Polygraph},
  author = {Vashurin, Roman and Fadeeva, Ekaterina and Vazhentsev, Artem and Rvanova, Lyudmila and Vasilev, Daniil and Tsvigun, Akim and Petrakov, Sergey and Xing, Rui and Sadallah, Abdelrahman and Grishchenkov, Kirill and Panchenko, Alexander and Baldwin, Timothy and Nakov, Preslav and Panov, Maxim and Shelmanov, Artem},
  journal = {Transactions of the Association for Computational Linguistics},
  volume = {13},
  pages = {220--248},
  year = {2025},
  doi = {10.1162/tacl_a_00737},
  url = {https://aclanthology.org/2025.tacl-1.11/}
}

@article{kuhn2023semantic,
  title = {Detecting Hallucinations in Large Language Models Using Semantic Entropy},
  author = {Farquhar, Sebastian and Kossen, Jannik and Kuhn, Lorenz and Gal, Yarin},
  journal = {Nature},
  volume = {630},
  pages = {625--630},
  year = {2024},
  doi = {10.1038/s41586-024-07421-0},
  url = {https://doi.org/10.1038/s41586-024-07421-0}
}

@inproceedings{khanmohammadi2026reliable,
  title = {How Reliable are Confidence Estimators for Large Reasoning Models? A Systematic Benchmark on High-Stakes Domains},
  author = {Khanmohammadi, Reza and Miahi, Erfan and Kaur, Simerjot and Smiley, Charese and Brugere, Ivan and Thind, Kundan S. and Ghassemi, Mohammad M.},
  booktitle = {Proceedings of the 19th Conference of the European Chapter of the Association for Computational Linguistics},
  pages = {1669--1754},
  year = {2026},
  publisher = {Association for Computational Linguistics},
  doi = {10.18653/v1/2026.eacl-long.78},
  url = {https://aclanthology.org/2026.eacl-long.78/}
}

@misc{ballon2026probing,
  title = {Probing the Trajectories of Reasoning Traces in Large Language Models},
  author = {Ballon, Marthe and Verbeken, Brecht and Ginis, Vincent and Algaba, Andres},
  note = {arXiv:2601.23163},
  year = {2026},
  doi = {10.48550/arXiv.2601.23163},
  url = {https://arxiv.org/abs/2601.23163}
}

@misc{wang2026cdg,
  title = {Inference Time Optimization with Confidence Dynamics},
  author = {Wang, Yu and Liu, Minghao and Wang, Jiayun and Huang, Jinrui and Shah, Ankit and Wei, Wei},
  note = {arXiv:2605.25244},
  year = {2026},
  doi = {10.48550/arXiv.2605.25244},
  url = {https://arxiv.org/abs/2605.25244}
}

@inproceedings{tikhonov2026confidence,
  title = {Confidence Leaps in {LLM} Reasoning: Early Stopping and Cross-Model Transfer},
  author = {Tikhonov, Pavel and Oseledets, Ivan and Tutubalina, Elena},
  booktitle = {Proceedings of the 19th Conference of the European Chapter of the Association for Computational Linguistics (Volume 2: Short Papers)},
  pages = {602--616},
  month = mar,
  year = {2026},
  address = {Rabat, Morocco},
  publisher = {Association for Computational Linguistics},
  doi = {10.18653/v1/2026.eacl-short.43},
  url = {https://aclanthology.org/2026.eacl-short.43/}
}

@misc{zhang2026commit,
  title = {When Does a Language Model Commit? A Finite-Answer Theory of Pre-Verbalization Commitment},
  author = {Zhang, Long and Chen, Wei-neng and Wei, Feng-feng and Qin, Zi-bo},
  note = {arXiv:2605.06723},
  year = {2026},
  doi = {10.48550/arXiv.2605.06723},
  url = {https://arxiv.org/abs/2605.06723}
}

@misc{gruenefeld2026tracing,
  title = {Tracing Uncertainty in Language Model ``Reasoning''},
  author = {Gr{\"u}nefeld, Nils and H{\o}jer, Bertram and Mondorf, Philipp and Plank, Barbara and Rogers, Anna and Hardmeier, Christian and Heinrich, Stefan and Frellsen, Jes},
  note = {arXiv:2605.07776},
  year = {2026},
  doi = {10.48550/arXiv.2605.07776},
  url = {https://arxiv.org/abs/2605.07776}
}

@misc{yang2025qwen3,
  title = {{Qwen3} Technical Report},
  author = {Yang, An and Li, Anfeng and Yang, Baosong and Zhang, Beichen and Hui, Binyuan and Zheng, Bo and Yu, Bowen and Gao, Chang and Huang, Chengen and Lv, Chenxu and others},
  note = {arXiv:2505.09388},
  year = {2025},
  doi = {10.48550/arXiv.2505.09388},
  url = {https://arxiv.org/abs/2505.09388}
}

@misc{abdin2024phi4,
  title = {{Phi-4} Technical Report},
  author = {Abdin, Marah and Aneja, Jyoti and Behl, Harkirat and Bubeck, S{\'e}bastien and Eldan, Ronen and Gunasekar, Suriya and Harrison, Michael and Hewett, Russell J. and Javaheripi, Mojan and Kauffmann, Piero and others},
  note = {arXiv:2412.08905},
  year = {2024},
  doi = {10.48550/arXiv.2412.08905},
  url = {https://arxiv.org/abs/2412.08905}
}

@misc{cobbe2021gsm8k,
  title = {Training Verifiers to Solve Math Word Problems},
  author = {Cobbe, Karl and Kosaraju, Vineet and Bavarian, Mohammad and Chen, Mark and Jun, Heewoo and Kaiser, Lukasz and Plappert, Matthias and Tworek, Jerry and Hilton, Jacob and Nakano, Reiichiro and Hesse, Christopher and Schulman, John},
  note = {arXiv:2110.14168},
  year = {2021},
  doi = {10.48550/arXiv.2110.14168},
  url = {https://arxiv.org/abs/2110.14168}
}

@inproceedings{hendrycks2021math,
  title = {Measuring Mathematical Problem Solving With the {MATH} Dataset},
  author = {Hendrycks, Dan and Burns, Collin and Kadavath, Saurav and Arora, Akul and Basart, Steven and Tang, Eric and Song, Dawn and Steinhardt, Jacob},
  booktitle = {Proceedings of the Neural Information Processing Systems Track on Datasets and Benchmarks},
  volume = {1},
  year = {2021},
  url = {https://datasets-benchmarks-proceedings.neurips.cc/paper/2021/hash/be83ab3ecd0db773eb2dc1b0a17836a1-Abstract-round2.html}
}

@inproceedings{lewkowycz2022minerva,
  title = {Solving Quantitative Reasoning Problems with Language Models},
  author = {Lewkowycz, Aitor and Andreassen, Anders and Dohan, David and Dyer, Ethan and Michalewski, Henryk and Ramasesh, Vinay and Slone, Ambrose and Anil, Cem and Schlag, Imanol and Gutman-Solo, Theo and others},
  booktitle = {Advances in Neural Information Processing Systems},
  volume = {35},
  year = {2022},
  url = {https://papers.nips.cc/paper_files/paper/2022/hash/18abbeef8cfe9203fdf9053c9c4fe191-Abstract-Conference.html}
}

@inproceedings{he2024olympiad,
  title = {{OlympiadBench}: A Challenging Benchmark for Promoting {AGI} with Olympiad-Level Bilingual Multimodal Scientific Problems},
  author = {He, Chaoqun and Luo, Renjie and Bai, Yuzhuo and Hu, Shengding and Thai, Zhen and Shen, Junhao and Hu, Jinyi and Han, Xu and Huang, Yujie and Zhang, Yuxiang and Liu, Jie and Qi, Lei and Liu, Zhiyuan and Sun, Maosong},
  booktitle = {Proceedings of the 62nd Annual Meeting of the Association for Computational Linguistics (Volume 1: Long Papers)},
  pages = {3828--3850},
  year = {2024},
  publisher = {Association for Computational Linguistics},
  doi = {10.18653/v1/2024.acl-long.211},
  url = {https://aclanthology.org/2024.acl-long.211/}
}

@misc{qwen2026qwen35,
  title = {{Qwen3.5-9B} Model Card},
  author = {{Qwen Team}},
  month = mar,
  year = {2026},
  url = {https://huggingface.co/Qwen/Qwen3.5-9B}
}

@misc{mistral2026ministral3,
  title = {Ministral 3},
  author = {Liu, Alexander H. and Khandelwal, Kartik and Subramanian, Sandeep and Jouault, Victor and Rastogi, Abhinav and others},
  note = {arXiv:2601.08584},
  year = {2026},
  doi = {10.48550/arXiv.2601.08584},
  url = {https://arxiv.org/abs/2601.08584}
}

@inproceedings{rein2024gpqa,
  title = {{GPQA}: A Graduate-Level Google-Proof {Q\&A} Benchmark},
  author = {Rein, David and Hou, Betty Li and Stickland, Asa Cooper and Petty, Jackson and Pang, Richard Yuanzhe and Dirani, Julien and Michael, Julian and Bowman, Samuel R.},
  booktitle = {First Conference on Language Modeling},
  year = {2024},
  url = {https://openreview.net/forum?id=Ti67584b98}
}

@inproceedings{suzgun2023bbh,
  title = {Challenging {BIG-Bench} Tasks and Whether Chain-of-Thought Can Solve Them},
  author = {Suzgun, Mirac and Scales, Nathan and Sch{\"a}rli, Nathanael and Gehrmann, Sebastian and Tay, Yi and Chung, Hyung Won and Chowdhery, Aakanksha and Le, Quoc V. and Chi, Ed H. and Zhou, Denny and Wei, Jason},
  booktitle = {Findings of the Association for Computational Linguistics: ACL 2023},
  month = jul,
  year = {2023},
  address = {Toronto, Canada},
  publisher = {Association for Computational Linguistics},
  pages = {13003--13051},
  doi = {10.18653/v1/2023.findings-acl.824},
  url = {https://aclanthology.org/2023.findings-acl.824/}
}

@inproceedings{wei2022cot,
  title = {Chain-of-Thought Prompting Elicits Reasoning in Large Language Models},
  author = {Wei, Jason and Wang, Xuezhi and Schuurmans, Dale and Bosma, Maarten and Ichter, Brian and Xia, Fei and Chi, Ed H. and Le, Quoc V. and Zhou, Denny},
  booktitle = {Advances in Neural Information Processing Systems},
  volume = {35},
  pages = {24824--24837},
  year = {2022},
  doi = {10.48550/arXiv.2201.11903}
}

@inproceedings{kojima2022zeroshot,
  title = {Large Language Models are Zero-Shot Reasoners},
  author = {Kojima, Takeshi and Gu, Shixiang Shane and Reid, Machel and Matsuo, Yutaka and Iwasawa, Yusuke},
  booktitle = {Advances in Neural Information Processing Systems},
  volume = {35},
  pages = {22199--22213},
  year = {2022},
  doi = {10.48550/arXiv.2205.11916}
}

@inproceedings{geifman2017selective,
  title = {Selective Classification for Deep Neural Networks},
  author = {Geifman, Yonatan and El-Yaniv, Ran},
  booktitle = {Advances in Neural Information Processing Systems},
  volume = {30},
  pages = {4878--4887},
  year = {2017},
  doi = {10.48550/arXiv.1705.08500}
}

@misc{aime24,
      title={American Invitational Mathematics Examination (AIME) 2024}, 
      author={Zhang, Yifan and Math-AI, Team},
      year={2024},
}

@misc{pandey2026selfdoubt,
  title = {{SELFDOUBT}: Uncertainty Quantification for Reasoning {LLM}s via the Hedge-to-Verify Ratio},
  author = {Pandey, Satwik and Raghu, Suresh and Pandey, Shashwat},
  note = {arXiv:2604.06389},
  year = {2026},
  doi = {10.48550/arXiv.2604.06389},
  url = {https://arxiv.org/abs/2604.06389}
}

@inproceedings{azaria2023internal,
  title={The internal state of an LLM knows when it’s lying},
  author={Azaria, Amos and Mitchell, Tom},
  booktitle={Findings of the Association for Computational Linguistics: EMNLP 2023},
  pages={967--976},
  year={2023}
}

@inproceedings{duan2024shifting,
  title={Shifting attention to relevance: Towards the predictive uncertainty quantification of free-form large language models},
  author={Duan, Jinhao and Cheng, Hao and Wang, Shiqi and Zavalny, Alex and Wang, Chenan and Xu, Renjing and Kailkhura, Bhavya and Xu, Kaidi},
  booktitle={Proceedings of the 62nd Annual Meeting of the Association for Computational Linguistics (Volume 1: Long Papers)},
  pages={5050--5063},
  year={2024}
}

@article{lin2023generating,
  title={Generating with confidence: Uncertainty quantification for black-box large language models},
  author={Lin, Zhen and Trivedi, Shubhendu and Sun, Jimeng},
  journal={arXiv preprint arXiv:2305.19187},
  year={2023}
}

@inproceedings{su2024unsupervised,
  title     = {Unsupervised Real-Time Hallucination Detection based on the Internal States of Large Language Models},
  author    = {Su, Weihang and Wang, Changyue and Ai, Qingyao and Hu, Yiran and Wu, Zhijing and Zhou, Yujia and Liu, Yiqun},
  booktitle = {Findings of the Association for Computational Linguistics: ACL 2024},
  pages     = {14379--14391},
  year      = {2024},
  publisher = {Association for Computational Linguistics},
  doi       = {10.18653/v1/2024.findings-acl.854},
  url       = {https://aclanthology.org/2024.findings-acl.854/}
}

\clearpage
\appendix
\section*{Appendix}

\section{Algorithm of \m}\label{app:alg}

\begin{algorithm}[ht]
\caption{\m Inference.}
\label{alg:trac}
\begin{algorithmic}[1]
\REQUIRE problem $x$, trace $r$, returned answer $a$, frozen LM $p_\theta$, trained head $g$
\ENSURE correctness score $u$
\STATE close $(x,r)$ with the fixed answer cue and reuse the cached prefix
\STATE decode $\tilde a$ and record token log-probabilities
\STATE $e\leftarrow\mathbb{I}[\tilde a\equiv a]$; compute $\ell,\ell_{\min},\ell_{\mathrm{head}},c_1$
\STATE $\phi_A\leftarrow[e;\ell;\ell_{\min};\ell_{\mathrm{head}};c_1]$ \COMMENT{\pcem}
\STATE $\phi_P\leftarrow\textsc{TraceProfile}(r)$ \COMMENT{\tupm}
\RETURN $u=g([\phi_A;\phi_P])$
\end{algorithmic}
\end{algorithm}

Algorithm~\ref{alg:trac} summarizes \m inference. Given a completed reasoning trace and its returned answer, \m appends a fixed answer cue to the cached prefix and decodes a short re-elicited answer. It then extracts agreement and token-level confidence features from this decoding, combines them with a passive profile of the original trace, and feeds the resulting representation into the trained head to produce a correctness score.

\section{Illustrative Example}\label{app:example}

We use a simple algebraic example to illustrate how \m identifies whether a returned answer is supported by its completed reasoning trace. Consider the problem
\begin{quote}
\textbf{Problem $x$:} Solve the equation $3(x-2)=24$.
\end{quote}
Suppose the frozen LLM generates the following reasoning body:
\begin{quote}
\textbf{Reasoning trace $r$:} Dividing both sides by $3$ gives $x-2=8$. Adding $2$ to both sides gives $x=10$.
\end{quote}
Although the trace correctly derives $x=10$, assume that the model mistakenly copies the intermediate value and returns
\begin{quote}
\textbf{Returned answer $a$:} $8$.
\end{quote}
This error can be difficult to identify from the returned-answer confidence alone because $8$ already appears as a valid intermediate result in the trace and may be generated with high token probability.

Following Algorithm~\ref{alg:trac}, \pcem reuses the cached representation of $(x,r)$, appends the fixed answer cue, and performs a short greedy decoding:
\begin{quote}
\textbf{Appended cue:} \texttt{$\langle$The final answer is$\rangle$}

\textbf{Re-elicited answer $\tilde a$:} $10$.
\end{quote}
The re-elicited answer is consistent with the conclusion derived by the reasoning trace but differs from the originally returned answer. Therefore, the agreement feature is
\begin{equation}
e=\mathbb{I}[\tilde a\equiv a]=\mathbb{I}[10\equiv 8]=0.
\end{equation}
Suppose the answer-only decoding strongly supports $\tilde a=10$ and produces the illustrative statistics
\begin{equation}
\ell=-0.06,\qquad \ell_{\min}=-0.11,\qquad \ell_{\mathrm{head}}=-0.07,\qquad c_1=-0.05.
\end{equation}
The active representation is consequently
\begin{equation}
\phi_A=[0;-0.06;-0.11;-0.07;-0.05].
\end{equation}
The low-magnitude log-probabilities indicate that the model confidently re-elicits $10$, while the zero agreement feature shows that this trace-supported answer is inconsistent with the returned answer $8$.

In parallel, \tupm summarizes the token-level uncertainty trajectory of the original response:
\begin{equation}
\phi_P=\textsc{TraceProfile}(r).
\end{equation}
Because the reasoning steps themselves are short, fluent, and locally confident, the passive profile may not provide a strong indication of failure. The active disagreement therefore supplies complementary evidence that cannot be obtained by inspecting the original token probabilities alone. After combining both views, the trained head may produce a low correctness score:
\begin{equation}
u=g([\phi_A;\phi_P])=0.08.
\end{equation}

For comparison, consider an otherwise identical response that correctly returns $a=10$. Re-elicitation again produces $\tilde a=10$, but the agreement feature becomes
\begin{equation}
e=\mathbb{I}[10\equiv 10]=1.
\end{equation}
With similar answer-level likelihood statistics and the same reasoning profile, the trained head may instead assign a high correctness score, such as
\begin{equation}
u=g([\phi_A;\phi_P])=0.94.
\end{equation}
This paired example highlights the role of trace-conditioned answer consistency. The two responses have the same problem and reasoning body, and both returned answers may appear locally plausible. However, re-elicitation directly tests which answer is supported by the completed reasoning context: it disagrees with the erroneous intermediate answer $8$ but reproduces the correct conclusion $10$. All confidence values and final scores in this example are illustrative.

\section{Reproducibility and Full Pipeline}\label{app:pipeline}
\subsection{Offline Preparation and Online Scoring}
For each problem, the frozen backbone produces one primary trace and, where sampling methods are evaluated, seven additional traces. A deterministic task-specific verifier labels the primary answer. Re-elicitation acts on the completed reasoning body, conditioning on the full prefix once without using correctness or the reference answer.
The completed prefix receives the fixed cue in Appendix~\ref{app:prompts}. The model greedily decodes a short answer and records its per-token log-probabilities. The re-elicited answer forms the agreement flag, its token log-likelihoods form the likelihood and confidence channels of $\phi_A$, and token-level uncertainty from the completed trace forms $\phi_P$. This is a single re-elicitation at the completed prefix. Fold-local standardization and a logistic head are fitted only on the training partition.
Online scoring re-elicits one cached short answer at the completed prefix, applies stored preprocessing, and evaluates Eq.~\eqref{eq:trac}. It uses no verifier or reference answer.

\subsection{Datasets, Models, and Protocol}\label{app:setup}
We evaluate our method on five mathematical datasets, including GSM8K~\cite{cobbe2021gsm8k}, MATH500~\cite{hendrycks2021math}, Minerva~\cite{lewkowycz2022minerva}, OlympiadBench~\cite{he2024olympiad}, and AIME~\cite{aime24}.
The non-mathematical extension uses BIG-Bench Hard tasks \cite{suzgun2023bbh} and GPQA-Diamond \cite{rein2024gpqa}. Mathematical answers use deterministic symbolic and numeric verification; multiple-choice answers use exact normalized option letters.

Primary traces are sampled at temperature 0.7 and top-$p$ 0.95. Learned estimators use five independently shuffled five-fold partitions with question-level separation. LODO, LOMO, and global-head studies fit preprocessing and classifiers on source pairs only. No target identity, accuracy, or normalization statistic enters the head. AUROC measures ranking; AURC measures selective risk and does not imply probability calibration.

\subsection{Baselines and Hyperparameters}
Mean log-probability averages selected-token log-probabilities. Self-certainty measures next-token distributional concentration \cite{kang2025selfcertainty}. DeepConf-bottom summarizes the least-confident local window \cite{fu2025deepconf}. P(True) uses a separate self-evaluation call \cite{kadavath2022mostly}. SelfDoubt scores hedging markers against self-verification markers in the trace \cite{pandey2026selfdoubt}. Answer convergence uses only the re-elicited answer-identity agreement. SC@$K$ uses the modal answer fraction among $K$ normalized samples \cite{wang2023selfconsistency}. SelfCheckGPT@$K$ scores consistency of the response against $K$ independent samples \cite{manakul2023selfcheckgpt}. All learned single-trace controls use the same folds and logistic capacity unless a classifier-capacity experiment states otherwise.

\begin{table}[h]
\centering
\small
\caption{Statistics of datasets. }
\setlength{\tabcolsep}{5pt}
\begin{tabular}{lccc}
\toprule
\textbf{Benchmark} & \textbf{Pairs} & \textbf{Problems} & \textbf{Regime} \\
\midrule
GSM8K & 6 & 796--800 & Grade-school word problems \\
MATH500 & 6 & 491--500 & Competition mathematics \\
Minerva & 6 & 264--272 & Technical quantitative \\
OlympiadBench & 6 & 579--581 & Olympiad-level problems \\
AIME & 6 & 60 & Competition mathematics \\
\bottomrule
\end{tabular}
\label{tab:app-datasets}
\end{table}

\begin{table}[t]
\centering
\small
\caption{Hyperparameter configuration.}
\setlength{\tabcolsep}{4pt}
\begin{tabular}{ll}
\toprule
\textbf{Parameter} & \textbf{Reported value} \\
\midrule
Full samples per problem & 8 \\
Sampling temperature / top-$p$ & $0.7$ / $0.95$ \\
Re-elicitation probes & 1 (completed prefix) \\
Probe decoding & Greedy, at most 16 tokens \\
\tupm position bins $J$ & 8 \\
\tupm entropy support $k$ & 5 \\
\tupm final-window size $W$ & 30 tokens \\
\tupm low-confidence threshold $\tau$ & $-2.0$ log-probability \\
\tupm auxiliary dimension $d_{\mathrm{aux}}$ & 11 \\
\tupm representation dimension $d_P$ & 27 \\
Classifier & Standardized logistic regression \\
Classifier penalty / $C$ & $\ell_2$ / $1.0$ \\
Classifier iterations & At most 1,000 \\
Evaluation folds & 5-fold stratified \\
Fold seeds & 2026, 7, 13, 42, 100 \\
Bootstrap & Stratified by correctness, per pair \\
\bottomrule
\end{tabular}
\label{tab:app-hyper}
\end{table}

\section{Interface Refinements and Learning Diagnostics}\label{app:measurement}

\begin{table}[t]
\centering
\small
\caption{\pcem design diagnostics.
\textbf{(a)} Adding each answer-interface feature to its matched base improves AUROC.
\textbf{(b)} Higher-capacity heads improve in-domain AUROC, but their gains over logistic regression shrink substantially and are
not consistent across classifiers under leave-one-dataset-out transfer. We therefore retain the logistic head for the main results.}
\setlength{\tabcolsep}{2pt}
\begin{tabular}{lrrrr}
\toprule

\multicolumn{5}{l}{\textbf{(a) Matched interface refinements}}\\
& \textbf{Base}
& \textbf{Variant}
& \multicolumn{2}{c}{$\boldsymbol{\Delta}$} \\
\midrule
$+$ Primary-answer agreement
& .720
& \textbf{.834}
& \multicolumn{2}{c}{$+.114$} \\

$+$ Full-answer likelihood
& .792
& \textbf{.834}
& \multicolumn{2}{c}{$+.042$} \\

\midrule
\multicolumn{5}{l}{\textbf{(b) Classifier-capacity control
(same \pcem features)}}\\
& \multicolumn{2}{c}{\textbf{In-domain}}
& \multicolumn{2}{c}{\textbf{LODO}} \\
\cmidrule(lr){2-3}\cmidrule(lr){4-5}
\textbf{Classifier}
& \textbf{AUROC}
& $\boldsymbol{\Delta}_{\mathrm{LR}}$
& \textbf{AUROC}
& $\boldsymbol{\Delta}_{\mathrm{LR}}$ \\
\midrule
Logistic regression
& .834
& reference
& .846
& reference \\

Random forest
& \textbf{.870}
& $+.036$
& \textbf{.862}
& $+.016$ \\

Gradient-boosted trees
& \underline{.855}
& $+.021$
& \underline{.851}
& $+.005$ \\

MLP
& .838
& $+.004$
& .845
& $-.001$ \\

\bottomrule
\end{tabular}
\label{tab:app-diagnostics}
\end{table}

Table~\ref{tab:app-diagnostics}(a) isolates the two answer-interface choices inside \pcem. Adding primary-to-re-elicited answer agreement to a confidence base raises AUROC substantially, and adding length-normalized full-answer likelihood on top of first-token confidence and agreement adds a further gain. Both refinements are incorporated into the \pcem module. Table~\ref{tab:app-diagnostics}(b) reports learning controls. Label efficiency is favorable: at 25\% of labels \pcem retains 95.9\% of full-data AUROC. A random forest improves in-domain results but retains only a marginal gain in leave-one-dataset-out transfer, so the primary evaluation uses the simpler logistic head.

\section{Complete and Additional Results}\label{app:results}
\subsection{Full Mathematical Matrix}

Table~\ref{tab:app-matrix} reports every pair. All 30 mathematical pairs enter the headline aggregates; pair-equal aggregation with a hierarchical bootstrap keeps any single small pair from dominating or destabilizing the macro.

\subsection{Consensus Blind Spot and Fusion}\label{app:fusion}
\begin{table}[t]
\centering
\small
\caption{AUROC on increasingly high-agreement subsets.}
\setlength{\tabcolsep}{9pt}
\begin{tabular}{rrrrrr}
\toprule
\textbf{Vote $\geq$} & \textbf{Pairs} & $n$ & \textbf{Wrong} & \textbf{SC} & \textbf{\pcem} \\
\midrule
.625 & 30 & 9,744 & 924 & \underline{.806} & \textbf{.895} \\
.750 & 30 & 9,036 & 574 & \underline{.729} & \textbf{.876} \\
.875 & 30 & 8,173 & 334 & \underline{.598} & \textbf{.851} \\
1.000 & 30 & 7,485 & 243 & \underline{.500} & \textbf{.839} \\
\bottomrule
\end{tabular}
\label{tab:app-blindspot}
\end{table}

As the vote threshold approaches unanimity, SC loses ranking resolution as shown in Table~\ref{tab:app-blindspot}. \pcem still ranks the 243 errors inside the unanimous subset. This finding explains why an active within-trace observation can complement agreement across completed traces.

\begin{table*}[t]
\centering
\small
\caption{Matched-budget fusion (\cfm vs.\ SC@8) by pair, for the four non-AIME mathematical datasets. $^{*}$ marks an uncorrected paired-bootstrap $p<.05$. The six AIME pairs are summarized in the last row and detailed in Table~\ref{tab:app-matrix}.}
\setlength{\tabcolsep}{12pt}
\begin{tabular}{ll cccc c}
\toprule
\textbf{Dataset} & \textbf{Model} & \textbf{SC@8} & \textbf{SemEnt@8} & \textbf{Conf.-vote} & \textbf{\cfm} & $\boldsymbol{\Delta}$ \\
\midrule
GSM8K & Ministral & .839 & .838 & .836 & .892 & $+.053^{*}$ \\
GSM8K & Phi-4-r & .820 & .820 & .828 & .856 & $+.036^{*}$ \\
GSM8K & Qwen3-14B & .811 & .812 & .811 & .887 & $+.076^{*}$ \\
GSM8K & Qwen3.5-9B & .908 & .918 & .904 & .953 & $+.045^{*}$ \\
GSM8K & Qwen3-4B & .894 & .892 & .894 & .896 & $+.002$ \\
GSM8K & Qwen3-8B & .823 & .823 & .823 & .931 & $+.108^{*}$ \\
MATH500 & Ministral & .914 & .913 & .914 & .969 & $+.055^{*}$ \\
MATH500 & Phi-4-r & .974 & .977 & .974 & .969 & $-.005$ \\
MATH500 & Qwen3-14B & .968 & .970 & .968 & .990 & $+.022^{*}$ \\
MATH500 & Qwen3.5-9B & .885 & .891 & .887 & .941 & $+.056^{*}$ \\
MATH500 & Qwen3-4B & .936 & .934 & .936 & .923 & $-.012$ \\
MATH500 & Qwen3-8B & .943 & .943 & .943 & .979 & $+.036^{*}$ \\
Minerva & Ministral & .755 & .753 & .759 & .779 & $+.025^{*}$ \\
Minerva & Phi-4-r & .731 & .733 & .736 & .781 & $+.050^{*}$ \\
Minerva & Qwen3-14B & .791 & .794 & .791 & .798 & $+.007$ \\
Minerva & Qwen3.5-9B & .702 & .700 & .701 & .790 & $+.089^{*}$ \\
Minerva & Qwen3-4B & .834 & .841 & .834 & .847 & $+.013$ \\
Minerva & Qwen3-8B & .774 & .780 & .774 & .771 & $-.003$ \\
Olympiad & Ministral & .921 & .914 & .920 & .953 & $+.032^{*}$ \\
Olympiad & Phi-4-r & .900 & .902 & .903 & .923 & $+.023^{*}$ \\
Olympiad & Qwen3-14B & .941 & .939 & .943 & .965 & $+.024^{*}$ \\
Olympiad & Qwen3.5-9B & .831 & .848 & .839 & .958 & $+.127^{*}$ \\
Olympiad & Qwen3-4B & .941 & .943 & .941 & .961 & $+.020^{*}$ \\
Olympiad & Qwen3-8B & .948 & .945 & .948 & .965 & $+.017^{*}$ \\
\midrule
\multicolumn{5}{r}{\textbf{Macro difference over these 24 non-AIME pairs}} & & \textbf{+.037} \\
\multicolumn{5}{r}{\textbf{Macro difference over all 30 headline pairs (incl.\ AIME)}} & & \textbf{+.038} \\
\bottomrule
\end{tabular}
\label{tab:app-fusion}
\end{table*}

Table~\ref{tab:app-fusion} exposes negative cases as well as gains. Six richer answer-distribution statistics trail SC@8, and confidence-weighted voting is approximately tied with SC@8. Thus, the improvement is not reproduced by a more elaborate vote representation.

\subsection{Robustness and Transfer}\label{app:robustness}
\begin{table}[H]
\centering
\small
\caption{AUROC over three independent generation seeds. AC denotes the answer-convergence baseline (re-elicited answer-identity agreement only).}
\label{tab:seeds}
\setlength{\tabcolsep}{6pt}
\begin{tabular}{llrrr}
\toprule
Dataset & Model & Log-prob. & AC & \pcem \\
\midrule
MATH500 & Qwen3-8B & $.691\pm.008$ & $.647\pm.019$ & $\mathbf{.791\pm.009}$ \\
MATH500 & Ministral & $.515\pm.010$ & $.604\pm.055$ & $\mathbf{.669\pm.035}$ \\
Olympiad & Qwen3-8B & $.739\pm.010$ & $.733\pm.014$ & $\mathbf{.835\pm.013}$ \\
Olympiad & Ministral & $.654\pm.015$ & $.669\pm.008$ & $\mathbf{.781\pm.017}$ \\
\bottomrule
\end{tabular}
\end{table}

Across three independent generations, \pcem is stable on the four audited pairs (Table~\ref{tab:seeds}). Three semantically equivalent probe cues produce only a small macro spread; cue ensembling adds no gain. Leave-one-dataset-out transfer matches or slightly exceeds the in-domain head and stays within noise of it, so the head does not overfit per-pair structure. A single global head without any per-pair fitting remains close under a leave-one-pair-out evaluation.
LOMO is mixed: transfer within the Qwen3 family gains, while transfer to Ministral and Qwen3.5 loses. These results support cross-dataset reuse, not a model-family-invariant calibration map.

\begin{table*}[t]
\centering
\small
\caption{Controls for trace length and difficulty.}
\setlength{\tabcolsep}{12pt}
\begin{tabular}{ll ccc cc}
\toprule
\textbf{Dataset} & \textbf{Model} & \textbf{LP+entropy+length} & \textbf{+ stability} & $\boldsymbol{\Delta}$
& \textbf{Within-length} & $r$(stability,length) \\
\midrule
GSM8K & Qwen3-4B & .902 & .912 & +.010 & .705 & $-.407$ \\
MATH500 & Qwen3-4B & .844 & .923 & +.079 & .814 & $-.521$ \\
Minerva & Qwen3-4B & .769 & .836 & +.067 & .725 & $-.562$ \\
Olympiad & Qwen3-4B & .895 & .949 & +.054 & .701 & $-.577$ \\
GSM8K & Qwen3-8B & .932 & .940 & +.008 & .722 & $-.384$ \\
MATH500 & Qwen3-8B & .884 & .965 & +.082 & .827 & $-.527$ \\
Minerva & Qwen3-8B & .756 & .768 & +.012 & .635 & $-.405$ \\
Olympiad & Qwen3-8B & .918 & .960 & +.042 & .705 & $-.628$ \\
GSM8K & Ministral & .780 & .891 & +.111 & .817 & $-.371$ \\
MATH500 & Ministral & .875 & .958 & +.083 & .777 & $-.471$ \\
Minerva & Ministral & .686 & .788 & +.102 & .699 & $-.263$ \\
Olympiad & Ministral & .887 & .946 & +.059 & .721 & $-.565$ \\
\midrule
\textbf{Macro} & & .844 & .903 & \textbf{+.059} & \textbf{.737} & $-.473$ \\
\bottomrule
\end{tabular}
\label{tab:app-length}
\end{table*}

The length control shows that active stability remains useful after coarse length matching. It does not establish independence from every notion of problem difficulty.

\begin{table}[H]
\centering
\small
\caption{Answer-leakage control. Traces are split by whether the reasoning body already states a string- or numerically-equivalent copy of the final answer. Even when the answer never appears in the trace, \m stays well above chance, so the signal is not explained by copying an answer already present in the text. The answer is copied into the reasoning body in roughly 78\% of traces overall. }
\setlength{\tabcolsep}{12pt}
\begin{tabular}{lrrr}
\toprule
\textbf{Subset} & \textbf{Pairs} & \textbf{\pcem} & \textbf{\m} \\
\midrule
Answer stated in trace & 30 & .851 & \textbf{.912} \\
Answer not stated in trace & 30 & .804 & \textbf{.859} \\
\bottomrule
\end{tabular}
\label{tab:app-leakage}
\end{table}

Table~\ref{tab:app-leakage} addresses whether \pcem merely copies an answer that already appears in the reasoning body. We split traces by whether the body contains a string- or numerically-equivalent copy of the final answer; the answer is present in roughly 78\% of traces. On the harder subset where the answer never appears, \m stays well above chance, so re-elicitation is not reducible to string copying, though the signal is naturally stronger when the answer has already been written into the trace.

\subsubsection{Non-Mathematical Reasoning}
\pcem improves over convergence plus endpoint confidence in-domain. Math-to-non-math transfer trails the best transferred baseline, so the cross-domain transfer claim is negative.

\section{Efficiency and Hardware}\label{app:efficiency}
The wall-clock audit reported in Table~\ref{tab:pareto-main} uses Qwen3-8B, one NVIDIA B200, bfloat16 weights, tensor-parallel size one, vLLM 0.19.1, automatic prefix caching, and 64 queries per serving profile. The single cached endpoint probe adds 34 milliseconds to the MATH500 median. Disabling cache reuse increases probe overhead by roughly four times, raising \m to $1.16\times$ mean relative latency.
Primary traces average approximately 1,807, 3,105, and 6,396 decoded tokens on GSM8K, MATH500, and OlympiadBench. The endpoint probe adds 3.8\%, 2.5\%, and 1.4\% decoded tokens, respectively. Token counts and latency are reported separately because scheduling, batching, cache management, and memory traffic prevent direct conversion. 

\section{Prompts, Parsing, and Stored Records}
\label{app:prompts}

\paragraph{Primary generation.}
All models receive the same instruction:
\texttt{Solve the following problem step by step. End your response with the final answer in $\cdots$}
We then apply the corresponding model-specific chat template, including the required role tokens and end-of-turn delimiters.

\paragraph{Prefix probing.}
For each selected reasoning prefix, we append the model-compatible end-of-thinking marker followed by
\texttt{$\langle$The final answer is$\rangle$}.
Decoding terminates when the matching closing brace, an end-of-sequence token, or a maximum of 16 additional tokens is generated.

\paragraph{Parsing.}
Equivalence between probe answers is determined using the same normalization procedure without access to the reference answer.
For each probe, we store the cut index, normalized reasoning depth, raw generated suffix, canonicalized answer or missing-answer state, first-token log-probability, and number of decoded tokens.

\section{Negative Findings and Limitations}\label{app:limitations}
In-domain conformal risk control works, but validity collapses under domain shift and robust variants obtain near-zero coverage. These findings are excluded from the central claims rather than hidden.
The method requires token scores and the ability to continue decoding from a completed prefix, so it does not directly apply to APIs that hide log-probabilities. Closing a prefix changes model behavior and does not prove that the visible trace caused the final answer. The score is not a correctness guarantee; deployment thresholds require representative validation and monitoring under shift.
In addition, \m may support abstention and escalation, but high-stakes use requires target-specific validation of thresholds and failure costs.

\begin{table*}[t]
\centering
\small
\caption{Complete results of AUROC on five datasets over six reasoning models from three families. \pcem is the active re-elicitation head and \m is the full estimator. }
\setlength{\tabcolsep}{12pt}
\begin{tabular}{ll rr cccc}
\toprule
\textbf{Dataset} & \textbf{Model} & $n$ & \textbf{Acc.} &
\textbf{\pcem} & \textbf{\m} & \textbf{SC@8} & $\boldsymbol{\Delta}$(full,active) \\
\midrule
\multirow{6}{*}{GSM8K}
& Ministral & 796 & .916 & .806 & \textbf{.881} & \underline{.839} & +.075 \\
& Phi-4-r & 800 & .964 & .713 & \textbf{.846} & \underline{.820} & +.133 \\
& Qwen3-14B & 800 & .965 & .490 & \textbf{.896} & \underline{.811} & +.406 \\
& Qwen3.5-9B & 800 & .896 & \underline{.912} & \textbf{.933} & .908 & +.021 \\
& Qwen3-4B & 800 & .945 & .805 & \underline{.889} & \textbf{.894} & +.084 \\
& Qwen3-8B & 800 & .950 & .718 & \textbf{.927} & \underline{.823} & +.209 \\
\midrule
\multirow{6}{*}{MATH500}
& Ministral & 491 & .617 & \underline{.928} & \textbf{.948} & .914 & +.020 \\
& Phi-4-r & 500 & .912 & .849 & \underline{.875} & \textbf{.974} & +.026 \\
& Qwen3-14B & 500 & .824 & .946 & \textbf{.971} & \underline{.968} & +.025 \\
& Qwen3.5-9B & 500 & .584 & .877 & \textbf{.913} & \underline{.885} & +.036 \\
& Qwen3-4B & 500 & .772 & .863 & \underline{.907} & \textbf{.936} & +.044 \\
& Qwen3-8B & 500 & .792 & \underline{.946} & \textbf{.968} & .943 & +.022 \\
\midrule
\multirow{6}{*}{Minerva}
& Ministral & 264 & .367 & .739 & \underline{.745} & \textbf{.755} & +.006 \\
& Phi-4-r & 272 & .460 & .602 & \textbf{.733} & \underline{.731} & +.131 \\
& Qwen3-14B & 272 & .504 & .593 & \underline{.730} & \textbf{.791} & +.137 \\
& Qwen3.5-9B & 272 & .335 & \underline{.773} & \textbf{.789} & .702 & +.016 \\
& Qwen3-4B & 272 & .397 & .773 & \underline{.818} & \textbf{.834} & +.045 \\
& Qwen3-8B & 272 & .478 & .693 & \underline{.745} & \textbf{.774} & +.052 \\
\midrule
\multirow{6}{*}{Olympiad}
& Ministral & 579 & .383 & \underline{.923} & \textbf{.928} & .921 & +.005 \\
& Phi-4-r & 581 & .707 & .832 & \textbf{.913} & \underline{.900} & +.081 \\
& Qwen3-14B & 581 & .656 & .909 & \textbf{.953} & \underline{.941} & +.044 \\
& Qwen3.5-9B & 581 & .358 & \underline{.932} & \textbf{.954} & .831 & +.022 \\
& Qwen3-4B & 581 & .627 & .894 & \textbf{.941} & \textbf{.941} & +.047 \\
& Qwen3-8B & 581 & .616 & .940 & \textbf{.952} & \underline{.948} & +.012 \\
\midrule
\multirow{6}{*}{AIME}
& Ministral & 60 & .200 & \underline{.941} & .922 & \textbf{.978} & $-.019$ \\
& Phi-4-r & 60 & .583 & .757 & \underline{.874} & \textbf{.925} & +.117 \\
& Qwen3-14B & 60 & .450 & .937 & \textbf{.979} & \underline{.964} & +.042 \\
& Qwen3.5-9B & 60 & .100 & \textbf{.994} & \textbf{.994} & .892 & $-.001$ \\
& Qwen3-4B & 60 & .417 & \textbf{.985} & \underline{.966} & .827 & $-.019$ \\
& Qwen3-8B & 60 & .350 & \underline{.945} & .927 & \textbf{.966} & $-.018$ \\
\bottomrule
\end{tabular}
\label{tab:app-matrix}
\end{table*}

\begin{table*}[t]
\centering
\small
\caption{Non-mathematical AUROC on pairs with at least 15 errors.}
\setlength{\tabcolsep}{10pt}
\begin{tabular}{llrrrrrrrr}
\toprule
\textbf{Task} & \textbf{Model} & $n$ & \textbf{Wrong} & \textbf{Mean LP} &
\textbf{Self-cert.} & \textbf{Conv.+final} & \textbf{\pcem} & \textbf{SC@8} \\
\midrule
GPQA-Diamond & Qwen3-8B & 198 & 97 & .684 & .687 & .682 & \underline{.768} & \textbf{.796} \\
GPQA-Diamond & Phi-4-r & 198 & 83 & .589 & .617 & .556 & \underline{.648} & \textbf{.745} \\
GPQA-Diamond & Ministral & 197 & 92 & .572 & .578 & \underline{.592} & .589 & \textbf{.753} \\
BBH date & Qwen3-8B & 250 & 22 & .740 & .726 & .660 & \underline{.775} & \textbf{.821} \\
BBH date & Ministral & 237 & 31 & .306 & .312 & .552 & \underline{.645} & \textbf{.841} \\
\midrule
\multicolumn{4}{r}{\textbf{Macro}} & .578 & .584 & .608 & \underline{.685} & \textbf{.791} \\
\bottomrule
\end{tabular}
\label{tab:app-nonmath}
\end{table*}
\end{document}